%% file: acl_latex.tex
\pdfoutput=1

\documentclass[11pt]{article}

\usepackage[preprint]{acl}

\usepackage{times}
\usepackage{latexsym}
\usepackage{svg}

\usepackage[T1]{fontenc}

\usepackage[utf8]{inputenc}

\usepackage{microtype}

\usepackage{inconsolata}

\usepackage{graphicx}
\usepackage{listings,xcolor}
\usepackage[most]{tcolorbox}
\usepackage{inconsolata}

\usepackage{enumitem}
\usepackage{booktabs}
\usepackage{float}

\newtcblisting[auto counter]{codelisting}[2][]{sharp corners, 
    fonttitle=\bfseries, colframe=gray, listing only, 
    listing options={basicstyle=\small\ttfamily, breaklines=true, breakautoindent=false, breakindent=0pt, escapeinside={(*@}{@*)}
}, 
    title=Prompt segment \thetcbcounter: #2, #1}

\usepackage{changepage}

\newcommand{\coedit}{CoEdIT-XL}
\newcommand{\LlamaH}{Llama 3 8B}
\newcommand{\LlamaS}{Llama 3 70B}
\newcommand{\mistral}{Mistral 7B}
\newcommand{\gptIVmini}{GPT-4o mini}
\newcommand{\gptIV}{GPT-4o}
\newcommand{\ppluie}{ParaPLUIE}

\newcommand{\dataset}{\texttt{ParaRev}}
\title{Identifying Reliable Evaluation Metrics for Scientific Text Revision}

\author{
  \textbf{Léane Jourdan\textsuperscript{1}},
  \textbf{Florian Boudin\textsuperscript{1,2}},
  \textbf{Nicolas Hernandez\textsuperscript{1}},
  \textbf{Richard Dufour\textsuperscript{1}},  
\\
  \textsuperscript{1}Nantes Université, École Centrale Nantes, CNRS, LS2N, UMR 6004, F-44000 Nantes, France
\\
  \textsuperscript{2}JFLI, CNRS, Nantes University, France
\\
  \small{
    \textbf{Correspondence:} \href{mailto:leane.jourdan@univ-nantes.fr}{leane.jourdan@univ-nantes.fr}
  }
}

\begin{document}
\maketitle
\begin{abstract}
Evaluating text revision in scientific writing remains a challenge, as traditional metrics such as ROUGE and BERTScore primarily focus on similarity rather than capturing meaningful improvements. In this work, we analyse and identify the limitations of these metrics and explore alternative evaluation methods that better align with human judgments. We first conduct a manual annotation study to assess the quality of different revisions. Then, we investigate reference-free evaluation metrics from related NLP domains. Additionally, we examine LLM-as-a-judge approaches, analysing their ability to assess revisions with and without a gold reference. Our results show that LLMs effectively assess instruction-following but struggle with correctness, while domain-specific metrics provide complementary insights. We find that a hybrid approach combining LLM-as-a-judge evaluation and task-specific metrics offers the most reliable assessment of revision quality. 
\end{abstract}

\section{Introduction}

\begin{figure}
\centering
  \includegraphics[width=\linewidth]{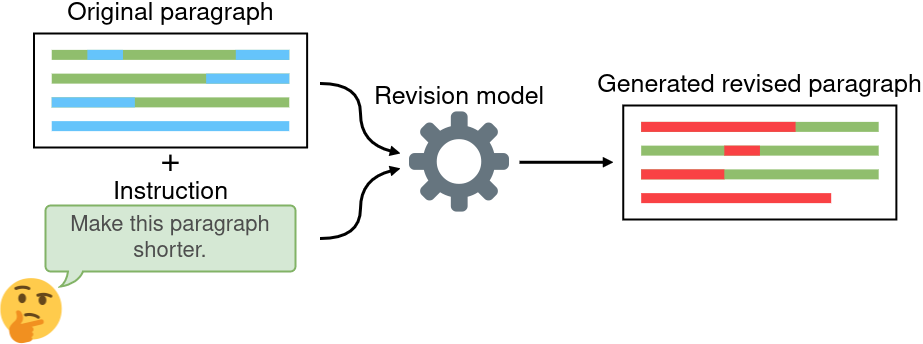}
  \caption{Overview of the text revision task}
  \label{fig:task_intro}
\end{figure}

Effective revision is a critical step in scientific writing, ensuring clarity, coherence, and adherence to academic standards. The writing process typically consists of four stages: 1) Prewriting, 2) Drafting, 3) Revising, and 4) Editing~\citep{jourdan2023text}. The revision stage involves substantial modifications to improve readability, style, and formality~\citep{du-etal-2022-read,li-etal-2022-text}. This step is particularly critical, as poor writing quality can obscure research findings and often contributes to paper rejection~\citep{amano2023manifold}.
As illustrated in Figure~\ref{fig:task_intro}, the revision task takes an original paragraph and an instruction specifying the required modification as input.
The expected output is a revised paragraph that aligns with the given instruction.

Given the importance of this task, reliable evaluation is crucial.  
Like other text generation tasks, text revision is assessed using well-established metrics such as ROUGE~\citep{lin-2004-rouge} or BERTScore~\citep{zhang2020bertscore}.
While embedding-based metrics (e.g., BERTScore) capture some semantic similarity, they still primarily focus on surface-level features and lexical overlap, rather than capturing deeper aspects of text quality.

In text revision, similarity-based metrics alone fail to fully capture revision quality. 
Beyond surface similarity to a reference text, revision assessment requires considering improvements over the original version, meaning preservation, and adherence to the instruction.
Several studies have relied on human evaluation to assess text revision systems~\citep{du-etal-2022-read,raheja2023coedit,raheja-etal-2024-medit,ito-etal-2020-langsmith,schick2023peer}. However, human evaluation is costly and time-consuming, making it impractical for large-scale or iterative assessments during system development.
To address this limitation, we explore alternative automatic evaluation approaches that provide a more reliable and scalable assessment of revision quality.

Since text revision encompasses various subtasks (e.g.,~paraphrasing, summarization, text simplification, style transfer, grammar error correction (GEC))~\citep{li-etal-2022-text,raheja-etal-2024-medit,ito-etal-2019-diamonds,kim-etal-2022-improving}, we first explore reference-free evaluation metrics commonly used to assess these tasks.
These metrics compare the original and revised texts directly, rather than relying on a gold reference.
Additionally, we explore different LLM-as-a-judge approaches, which can incorporate the revision instruction and potentially approximate human reasoning. 
With the rapid growth of LLMs, these approaches are increasingly used for evaluating diverse tasks~\citep{gu2024survey}.
However, prior studies have shown that LLMs experience performance drops when no gold reference is provided, sometimes being outperformed by simpler methods, making them less appealing~\citep{doostmohammadi-etal-2024-reliable,mita-etal-2024-towards}.
In this work, we aim to evaluate whether these results generalise to the text revision task and investigate the impact of providing a gold reference.
Our contributions are as follows:
\begin{itemize}[topsep=.5em, itemsep=.1em,leftmargin=1em]
    \item We release ParaReval, a dataset of human pairwise evaluations of generated revisions.\footnote{\href{https://github.com/JourdanL/parareval}{https://github.com/JourdanL/parareval}}
    \item We show that traditional similarity metrics fail to accurately evaluate text revision.
    \item We demonstrate that LLM-as-a-judge can effectively assess instruction following without the need for a gold reference.
    \item We find that similarity metrics complement LLM-as-a-judge in addressing challenging cases.
    \item While LLM-as-a-judge performs best, we show that the \ppluie\ metric~\cite{lemesle-etal-2025-paraphrase} can serve as a cost-effective alternative for measuring meaning preservation.
\end{itemize}

\section{Related Work}

In this section, we categorise evaluation approaches into three types: n-gram similarity metrics, embedding-based similarity metrics, and LLM-as-a-judge methods. 

\subsection{N-grams Similarity Metrics}

N-gram-based similarity metrics have been the standard for evaluating text generation tasks.
These metrics primarily measure lexical overlap between the generated text and the reference. However, they cannot capture semantic equivalence or improvements made over the original text.
The most commonly used n-gram-based metrics are:
\begin{itemize}[topsep=.5em, itemsep=.1em,leftmargin=1em]
    \item \textbf{BLEU}~\citep{papineni-etal-2002-bleu}: Initially developed for machine translation evaluation, BLEU has been widely used in text revision tasks~\citep{du-etal-2022-read,raheja-etal-2024-medit,jourdan-etal-2024-casimir,dwivedi-yu-etal-2024-editeval,mucke-2023}. 
    \item \textbf{ROUGE}~\citep{lin-2004-rouge}: Designed for summarization evaluation, ROUGE includes several variants, with \textbf{ROUGE-L} being the most commonly used in text revision~\citep{du-etal-2022-read,jourdan-etal-2024-casimir,jourdan-etal-2025-pararev,dwivedi-yu-etal-2024-editeval}. 
    \item \textbf{METEOR}~\citep{banerjee-lavie-2005-meteor}: A unigram matching metric for machine translation, less sensitive to paraphrasing than BLEU. Used in text revision by~\citet{mucke-2023}.
    \item \textbf{GLEU}~\citep{napoles-EtAl:2015:ACL-IJCNLP}: A variant of BLEU tailored for GEC, and used for text revision in~\citep{dwivedi-yu-etal-2024-editeval,raheja2023coedit}. 
    It takes the source text into account, rewarding corrections and crediting unchanged parts, while also penalizing ungrammatical edits.
    \item \textbf{SARI}~\citep{xu-etal-2016-optimizing}: Designed for automatic text simplification, it is the most commonly used metric for evaluating text revision~\citep{du-etal-2022-read,raheja2023coedit,raheja-etal-2024-medit,jourdan-etal-2024-casimir,jourdan-etal-2025-pararev,dwivedi-yu-etal-2024-editeval}. 
    It compares the system's output to both the reference and the source texts, rewarding correct additions, deletions, and retention of words.
    
\end{itemize}

BLEU, ROUGE and METEOR metrics score the similarity between the generated output and a reference text. SARI and GLEU are the only metrics that consider the source text, which is essential for assessing improvements in text revision. 
However, while they are interpretable, they struggle with tasks requiring deeper semantic understanding, such as evaluating instruction following or measuring improvements over the original text.

\subsection{Embedding Similarity Metrics}

Embedding-based metrics like BERTScore~\citep{zhang2020bertscore}, MoverScore~\citep{zhao-etal-2019-moverscore} or SemDist~\citep{kim21e_interspeech} are designed to capture semantic similarity beyond surface-level lexical overlap. 
These methods compare the embeddings of generated and reference texts to assess their alignment.
In text revision, only BERTScore has been used~\citep{mucke-2023,jourdan-etal-2024-casimir}.
It computes cosine similarity between contextualised embeddings from BERT for corresponding words in reference and generated sentences.

\subsection{LLM-as-a-Judge Approaches}

Recent works have explored the use of LLM-as-a-judge for evaluating generation tasks to go beyond surface similarity. These approaches treat evaluation as a judgment task, where an LLM assesses generated text based on multiple criteria. Several classification schemes have been proposed:
\citet{gu2024survey} propose to categorise them into \emph{Scores}, \emph{Yes or No}, \emph{Pair} and \emph{Multiple choice}.
\citet{zheng2023judgingllmasajudgemtbenchchatbot} propose three different variations: \emph{pairwise comparison}, \emph{single-answer grading}(score) and \emph{reference-guided grading}.

Notably,~\citet{doostmohammadi-etal-2024-reliable} proposed evaluating generated text on three dimensions: \emph{naturalness} (does the generation sound natural and fluent?), \emph{relatedness} (is the generation related to the prompt and follow the required format?), and \emph{correctness} (is the generation correct?, with meaning varying by task).
For text revision evaluation,~\citet{mita-etal-2024-towards} designed a one-question pairwise comparison prompt and tested it in a zero and few-shot settings. However, their results showed that this approach underperformed compared to a fine-tuned BERT classifier.

For our LLM-as-a-judge approaches, we build on these works and we propose a combination of the three variations from~\citep{zheng2023judgingllmasajudgemtbenchchatbot}.%

\section{Experimental Setup}

To examine the limitations of traditional similarity metrics, we first generate multiple revised outputs using various LLMs and manually evaluate them.

\subsection{Dataset}
\label{sec:gen_revisions}

We use the evaluation split of the \texttt{\dataset{}} dataset~\cite{jourdan-etal-2025-pararev}, which contains 258 pairs of revised paragraphs extracted from scientific articles revised by the authors themselves. Each paragraph is annotated with two different revision instructions, resulting in a total of 516 data points. Additionally, each paragraph is labelled with its revision intention type, which will be used in our analysis, 
the taxonomy is provided in Table~\ref{tab:taxonomy}.

\begin{table}[htb!]
  \centering
  \resizebox{\linewidth}{!}{
    \input{acl_tables/taxonomy-pararev}
    }
  \caption{Taxonomy of revisions at paragraph level}
  \label{tab:taxonomy}
\end{table}

\subsection{Revision Models}

To ensure a diverse set of revision outputs in terms of quality, we generate revised paragraphs for each \textit{original paragraph + instruction} pair using 6 different models. The models used are the following:
\begin{itemize}[topsep=.5em, itemsep=.1em,leftmargin=1em]
    %
    \item \href{https://huggingface.co/grammarly/coedit-xl}{\textbf{\coedit}}, a T5-based model fine-tuned for sentence revision~\citep{raheja2023coedit}
    \item Open LLMs: \textbf{\href{https://huggingface.co/meta-llama/Meta-Llama-3-8B-Instruct}{Llama 3 8B Instruct}}, \textbf{\href{https://huggingface.co/meta-llama/Meta-Llama-3-70B-Instruct}{Llama 3 70B Instruct}}, \textbf{\href{https://huggingface.co/mistralai/Mistral-7B-Instruct-v0.2}{Mistral 7B Instruct v0.2}}
    \item Closed-source LLMs: \textbf{GPT 4o mini}, \textbf{GPT 4o}
\end{itemize}
The prompts are provided in Appendix~\ref{apx:prompts_rev}.

\subsection{Annotation Task}

To identify which metrics best reflect the true quality of revisions, we conducted a manual evaluation comparing human judgments with automatic metric scores. For this, we designed an annotation task where human annotators compared pairs of revision candidates and selected their preferred version.

We carried out the annotation with the help of 10 annotators: 3 professors and 7 PhD students, all non-native English speakers but experienced in both reading and writing scientific papers in NLP.
Each annotation instance consisted of two revision suggestions for a given paragraph, produced in Section~\ref{sec:gen_revisions}, along with the corresponding revision instruction. 
Annotators answered a series of questions to assess the quality of the revisions:
\begin{itemize}[topsep=.5em, itemsep=.1em,leftmargin=1em]
    \item Q1A and Q1B \textbf{Relatedness} x2: \textit{Did model A/B address the instruction? \{Yes strictly, Yes with additional modifications, No\}}
    \item If it was your article and your instruction:
    \begin{itemize}[topsep=0em, itemsep=.1em,leftmargin=1em]
        \item Q2 \textbf{Correctness}: \textit{Which revisions would you consider acceptable? \{Both, A only, B only, None\}}
    \item Q3 \textbf{Preference}: \textit{Which revision would you prefer to include in your paper? \{Both, A, B, None\}}

    \end{itemize}
    \item Category-Specific Evaluation, \textit{\{Both, A, B, None\}} are possible answers for each question: 
     \begin{itemize}[topsep=0em, itemsep=.1em,leftmargin=1em]
         \item Rewriting light: \textit{Which model improves the academic style and English the most?}
    
         \item Rewriting medium: \textit{Which model improves the readability and structure the most?}
         \item Rewriting heavy: \textit{Which model improves the readability and clarity the most? }
    
         \item Concision: \textit{Which model manages the most to give a shorter version while keeping all the important ideas? }
     \end{itemize}
\end{itemize}
A screenshot of the annotation environment is available in Appendix~\ref{apx:annot_env}. To ensure fair evaluation, we balanced the pairwise comparisons across models, ensuring that each model was compared to the others a similar number of times. 

\subsection{Annotation Phase Results}

From the evaluation subset of the dataset, we generated 1,548 pairs of revised paragraphs for annotation. Among these, 129 pairs (8.33\%) received double annotations to measure inter-annotator agreement. The agreement scores (Cohen's Kappa $\kappa$) for each question are reported in Table~\ref{tab:agreement}.

For our analysis, as we are studying the metrics' capacity to identify the best revision among two propositions, we introduce the notion of Extended Preference. Even if annotators select \emph{None} for the Preference question, a model is still considered preferable if it is the only one \emph{Correct} or \emph{Related} to the instruction. We then consider the leftovers \emph{Both} and \emph{None} as \emph{Ties}.

\begin{table}[htb!]
  \centering
  \input{acl_tables/agreement}

  \caption{Cohen's Kappa ($\kappa$) for each question.}
  \label{tab:agreement}
\end{table}

\begin{figure}[htb!]
\centering
  \includegraphics[width=\columnwidth]{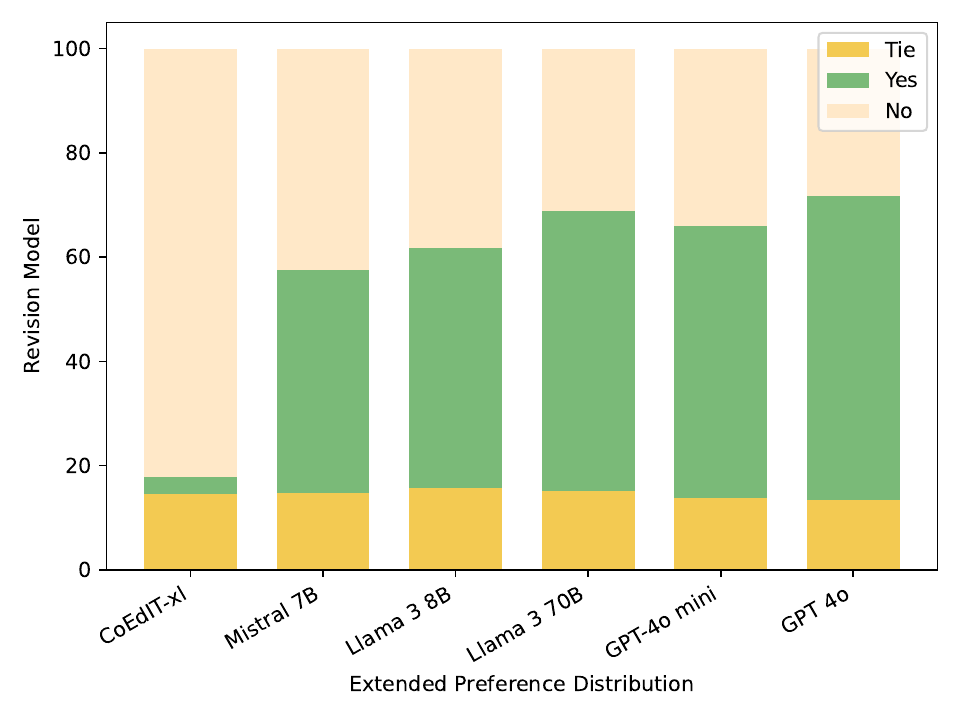}
  \caption{Distribution of human extended preference for each revision model. The green area indicates cases where the model is preferred.}
  \label{fig:human_pref_distrib}
\end{figure}

Figure~\ref{fig:human_pref_distrib} presents the distribution of human preferences across models. Based on human annotations, \gptIV\ emerges as the best-performing revision model, being strictly favoured in 58.33\% of comparisons. \LlamaS\ follows, with a preference rate of 53.68\%. When doing pairwise comparison of revision models, we observe that \LlamaS\ is in a tie with \gptIV . For more details, we also report the pairwise preferences on revision models in Appendix~\ref{apx:human_pref_matrix}.

\section{Limitations of Similarity-based Metrics}
In this section, we evaluate generated revisions and study the weaknesses of similarity-based metrics.
\subsection{Performance of Revision Models with Similarity-based Metrics}

To determine the best revision model, we evaluate each generated revision using traditional similarity-based metrics by comparing it to a reference.
Additionally, we compare the scores of these models with a no-edits baseline that simply recopies the input as output.
Results are presented in Table~\ref{tab:results-rev}.
We observe that all metrics, except GLEU, consider the no edits baseline to be the best-performing approach, with \coedit{} also being a strong contender. 
However, upon manual inspection, we find that \coedit{} tends to perform minimal revisions, such as correcting grammar and typos or, in some cases, excessively deleting parts of the paragraph. 
This suggests that these metrics favour not making any changes rather than rewarding meaningful, in-depth revisions.

\begin{table*}
    \centering
  \resizebox{\linewidth}{!}{
  \input{acl_tables/results_all}
  }
  \caption{Results on the paragraph revision task with traditional and alternative metrics.}
  \label{tab:results-rev}
\end{table*}

\subsection{Redundancy and Correlation Among Metrics}

\begin{figure}[htb!]
  \includegraphics[width=\columnwidth]{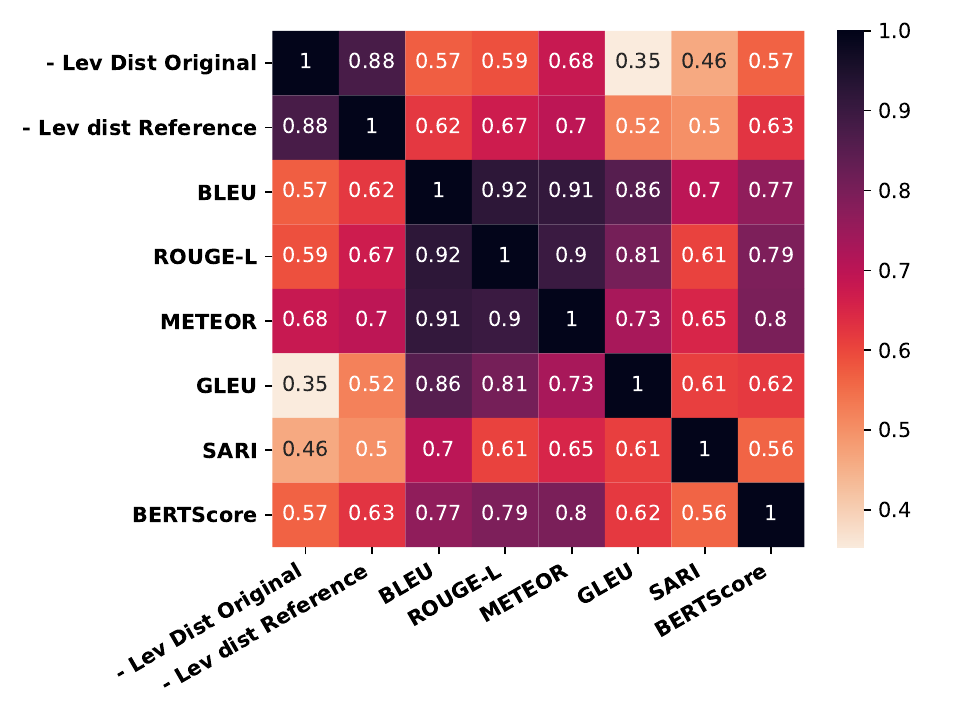}
  \caption{Correlation of the similarity metrics against each other and to the negative edit distance.}
  \label{fig:corr_old_metrics}
\end{figure}

To further investigate this issue, we analyse the correlation between different metrics and their relationship with edit distance (Levenshtein distance).
We compute the Pearson correlation coefficient between all metrics (see Figure~\ref{fig:corr_old_metrics}). 
We observe that most metrics are highly correlated, suggesting they provide redundant information.
The only exception is SARI, which differs from most other metrics because it considers the original text, the generated revision, and the reference.
These results suggest that, although we aim to use different metrics to study the revisions from various angles, most of them ultimately convey the same information.

\subsection{Sensitivity of Metrics to Edit Distance}

The first two columns of Figure~\ref{fig:corr_old_metrics} show a strong correlation between similarity metrics and negative edit distance, both in relation to the original and the reference paragraph. This relation is further illustrated in Appendix~\ref{apx:lev}. 
Two key observations emerge:
\begin{itemize}[topsep=.5em, itemsep=.1em,leftmargin=1em]
    \item \textbf{First, metrics only capture surface similarity}. The high correlation with the distance between the reference and generated revision suggests that traditional metrics mostly reflect how closely a model replicates the reference revision, rather than evaluating the quality of the revision itself. Even BERTScore, despite being based on embeddings and computationally more expensive, ultimately provides similar information to simpler distance-based metrics.
    \item \textbf{Second, substantial revisions are penalised}. The strong correlation between metric scores and the distance between the original and generated text indicates that the more a revision deviates from the original paragraph, the lower its score. This suggests that traditional metrics do not reward substantial, qualitative improvements, such as restructuring sentences or enhancing clarity. Instead, they encourage conservative edits that closely match the reference.
\end{itemize}

\begin{figure*}[htb!]
  \includegraphics[width=\textwidth]{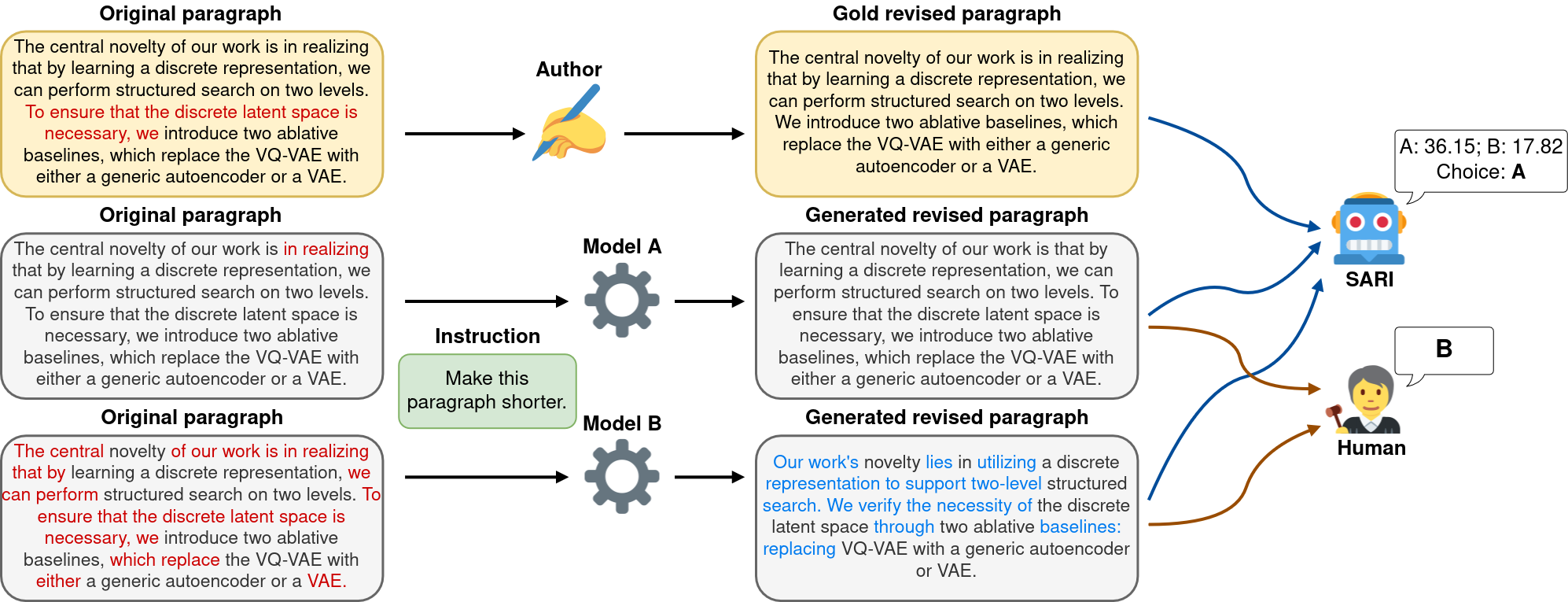}
  \caption{Overview of the evaluation in a case where automatic evaluation (here SARI) and human judgment don't align.}
  \label{fig:example_failure_main}
\end{figure*}

This phenomenon creates a major evaluation bias: Models that produce minimal edits receive higher scores and valid, but different, improvements are penalised. In many cases, making no revision at all results in a higher score than making meaningful changes, as exemplified in Figure~\ref{fig:example_failure_main}.

Among all metrics, SARI and GLEU stand out by having a lower correlation to edit distance ($\leq0.52$), as they explicitly penalise unchanged text edits, thereby encouraging revision.

\section{Exploring Alternative Evaluation Approaches} 
The goal of this section is to identify evaluation metrics that better correlate with human assessments of text revision quality.

\subsection{Metrics from Related NLP Domains} 
\label{sec:other_domains}

We hypothesise that in text revision, an essential factor is the comparison to the original text, as the metrics computing the similarity to a reference revision tend to overlook whether the modification effectively improves the original text. SARI and GLEU are widely used in text revision because they consider both the original and reference texts.

Additionally, text revision encompasses various subtasks depending on the type of modification conducted.~\citet{raheja-etal-2024-medit} classified revisions into three main categories: GEC, Simplification, and Paraphrasing. They evaluated each with a distinct set of metrics. This suggests that a single metric may not be sufficient for text revision, as we are not trying to capture the same phenomenon depending on the type of revision.

Inspired by these ideas, we explore metrics from related NLP domains, selecting those that consider the original text and align with specific types of revision (e.g., text summarization metrics for concision tasks or paraphrase evaluation metrics for rewriting tasks). 
We identify three candidate metrics, taking the original and generated revised paragraph as input:
\begin{itemize}[topsep=.5em, itemsep=.1em,leftmargin=1em]
    \item \textbf{BETS}~\cite{zhao-etal-2023-towards}: Designed for text simplification to assess meaning preservation and comparative simplicity at the level of modified word pairs, using BERT embeddings. 
    \item \textbf{BLANC}~\cite{vasilyev-etal-2020-fill} (BLANC-help variant): Designed for document summarization as a replacement for ROUGE. It measures how \emph{helpful} a summary is to understand a text, using a BERT-based model. 

    \item \textbf{\ppluie}~\cite{lemesle-etal-2025-paraphrase}: A metric for paraphrase detection that 
    prompts \mistral\ and uses perplexity scores when suggesting a \texttt{Yes} or \texttt{No} answer- instead of the generated text.
    %
\end{itemize}

Table~\ref{tab:results-rev} presents the evaluation results using these three candidate metrics, we also add the results comparing the original paragraph to the gold revision for comparison. BETS and \ppluie\ present a similar ranking of their preferred models and rank \coedit\ last like human annotation. Conversely, BLANC follows the one of similarity-based metrics.

\begin{figure}[ht!]
  \includegraphics[width=0.92\columnwidth]{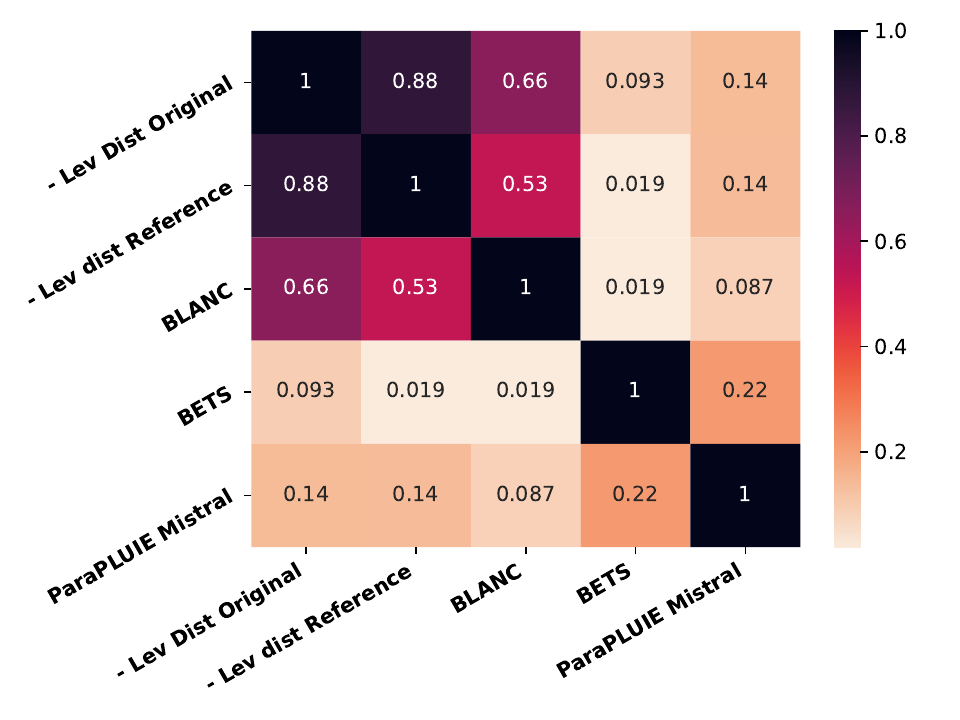}
  \caption{Correlation of the out-of-domain metrics against each other and to the negative edit distance.}
  \label{fig:corr_new_metrics}
\end{figure}

Figure~\ref{fig:corr_new_metrics} reports the correlations between these new metrics and negative edit distance. Except for BLANC, these metrics showcase a low correlation to the edit distance. Additionally, in Appendix~\ref{apx:lev}, we visualise these correlation relations.

BETS and \ppluie\ emerge as promising candidates for evaluating text revision, while BLANC appears less suitable. We further investigate their alignment with human annotations to confirm their effectiveness in Sections \ref{sec:results} and \ref{sec:analysis}.

\subsection{LLM-as-a-Judge}

An additional hypothesis is that text revision should not only account for the original text but also assess the model’s ability to follow instructions effectively. We explore LLM-as-a-Judge approaches to evaluate both these aspects.

We experiment with different approaches for LLM-as-a-judge, based on the work of~\citet{doostmohammadi-etal-2024-reliable}, who employed GPT-4 as a judge to evaluate three key criteria in generated text: 1) \emph{Naturalness}, 2) \emph{Relatedness}, and 3) \emph{Correctness}.
Since our task involves modifying existing text rather than generating it from scratch, \emph{Naturalness} is not relevant to our evaluation.
However, \emph{Relatedness} (whether the revision follows the instruction), aligns with Q1A and Q1B from our human annotation and \emph{Correctness} (whether the revision is acceptable) aligns directly with Q2. 
We structure our prompt similarly to the human annotation task to evaluate these two aspects.

We explore two approaches from~\citet{gu2024survey} for using LLMs as judges: Generating scores (\emph{LLM-Likert}) where the model is presented individual revisions to grade them, and Yes or No questions + Pairwise comparisons (\emph{LLM-Choice}) where the model is presented with pairs of revisions to select its preferred one or declare a tie. 
As~\citet{doostmohammadi-etal-2024-reliable} pointed out a drop in performance when \gptIV\ was not provided a gold reference, we experimented with these two approaches with and without a gold reference.
The prompts are provided in Appendix~\ref{apx:prompts_judge}.

Since our revision candidates were generated by multiple LLMs, we ensure a fair evaluation by also using multiple LLMs as judges.
This helps to reduce potential bias, where a model might favour its own outputs.
In Appendix~\ref{apx:bias}, we analyse the preferences of each LLM judge and discuss this potential bias.

We averaged agreements over three runs for all models except \gptIV\ due to cost constraints.
However, in the main body of the paper, we further average the results of all models per approach  to present them more concisely.

The results with LLM-as-a judge approaches are reported in Table~\ref{tab:results-llm}. All approaches agree with human annotation on \coedit\ being the less performing revision model and on LLama70B being a strong option for the task. However, LLM-likert scores are really close.

\begin{table}[htb!]
  \centering
  \resizebox{\linewidth}{!}{
  \input{acl_tables/results-llm}
  }
  \caption{Results on the paragraph revision task with LLM-as-a-judge evaluation. For LLM-choice, we report the extended preference and for LLM-likert, we report the average of all criteria.}
  \label{tab:results-llm}
\end{table}

\subsection{Results}
\label{sec:results}
After identifying all the candidate metrics, we assess their alignment with human judgement using three distinct measures to convey this agreement: Cohen's Kappa ($\kappa$), Cramér's V ($V$), and Pairwise Accuracy to account for ties~\citep{deutsch2023ties}.
Table~\ref{tab:overall_results} reports the alignment of automatic evaluation methods with human judgments, we add a random baseline for comparison.
As BLEU, ROUGE-L and METEOR are highly correlated, we report only ROUGE-L results in the main body of this paper.

LLM-Choice emerges as the most reliable evaluation option, followed by \ppluie. LLM-Likert and GLEU also exhibit strong alignment.
However, while LLM-Likert achieves higher accuracy when making a decision, it tends to overclassify cases as ties (See Appendix~\ref{sec:distrib_pref_llm_judge}).

\begin{table}[htb!]
  \centering
  \resizebox{\linewidth}{!}{
    \input{acl_tables/overall_results}
    }
  \caption{Alignment of automatic metrics with human judgements across all data. Pairwise accuracy and Cramér's V are defined on [0:1] and Cohen's Kappa on [-1:1].}
  \label{tab:overall_results}
\end{table}

\section{Performance by Aspects}
\label{sec:analysis}
In this section, we further analyse performance at a finer granularity on two aspects to see if the performances vary with the type of revision or the difficulty to discriminate the pair of propositions.

\subsection{Performance by Revision Category}

\begin{figure*}[t]
  \includegraphics[trim={0.1cm 0.6cm 0.3cm 0cm},width=\textwidth]{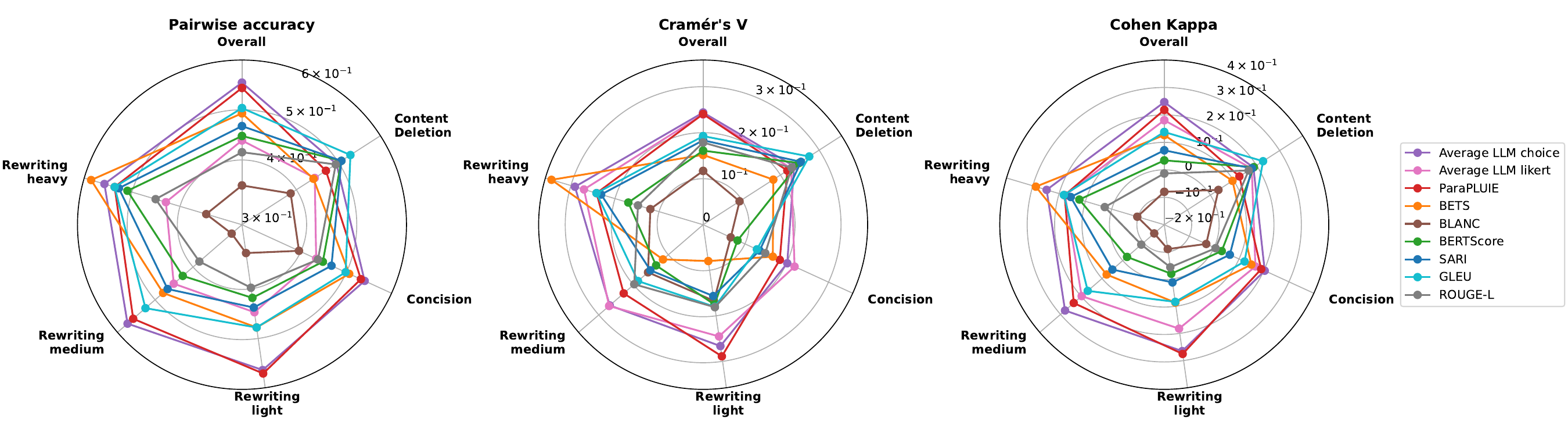}
  \caption{Alignment of automatic metrics with human annotations by revision category}
  \label{fig:alignement_categories}
\end{figure*}

To test our hypothesis from Section~\ref{sec:other_domains}, we analyse the alignment of human and automatic judgments across different revision types using \dataset\ label annotations (see~Figure~\ref{fig:alignement_categories}).
For most categories, LLM-Choice is the most reliable evaluation approach.
However, for cases where paragraph content is minimally altered (\texttt{Rewriting Light}, \texttt{Rewriting Medium} and \texttt{Concision}), \ppluie\ appears as a good alternative to capture meaning-preservation, as it is less costly than LLM-as-a-judge approaches. 
For instance, it processed our dataset in just 11 minutes, compared to 1 hour and 22 minutes required for \mistral-Choice on a V100 GPU. 

For \texttt{Content Deletion}, n-gram similarity-based metrics such as GLEU and SARI offer cost-efficient alternatives, aligning as well as or better than LLM-Choice with human preferences by leveraging reference-based deletion information.

Finally, for \texttt{Rewriting~Heavy}, BETS outperforms other metrics and aligns better to human annotation than in other categories of revisions.
In this category, the meaning of the paragraph must remain the same, while undergoing in-depth restructuring and rephrasing of most of the content.
In the dataset, many of the instructions associated with these paragraphs focus on making them clearer, more readable, or easier to understand, which can be linked to the task of text simplification. 
BETS is a balanced score between meaning preservation and text simplification, which likely explains its strong performance in this category.

\subsection{Performance by Difficulty}

To further assess metric effectiveness, we analyse performance across varying difficulty levels. We categorise revision pairs based on human annotation difficulty levels: 
\begin{itemize}[topsep=.5em, itemsep=.1em,leftmargin=1em]
    \item Easy Cases (530 pairs): defined by Q1A and Q1B, one model followed the instruction while the other did not.
    \item Medium Cases (214 pairs): defined by Q2, both models followed the instruction, but only one produced an acceptable revision.
    \item Hard Cases (575 pairs): defined by Q3: Both revisions were acceptable, but one was preferred.
\end{itemize}

\begin{figure*}
  \includegraphics[trim={0.1cm 0.6cm 0.3cm 0cm},width=\textwidth]{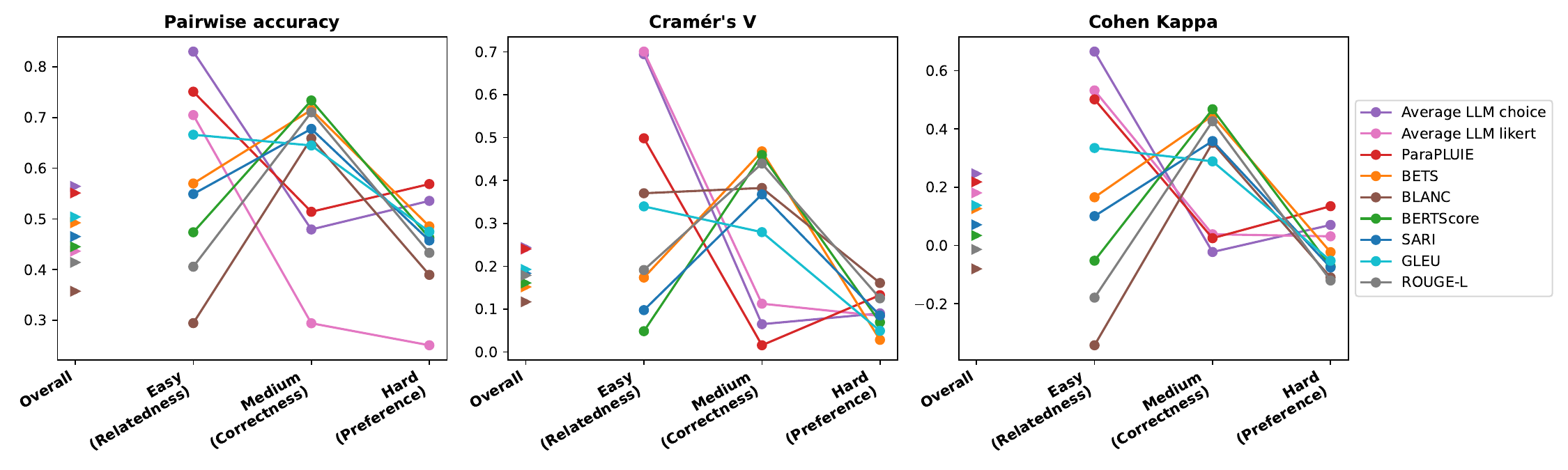}
  \caption{Alignment of automatic metrics with human annotations, by difficulty. The triangles in the first column represent the agreement on the total dataset from Table~\ref{tab:overall_results}}
  \label{fig:alignement_difficulty}
\end{figure*}

We report the alignment by difficulty in Figure~\ref{fig:alignement_difficulty}.
We observe that \textit{LLM-Choice} performs best in easy cases, achieving 0.821 accuracy. 
This suggests that LLMs are particularly effective at recognising whether a revision follows the given instruction, an ability that none of the other metrics possess. 
However, for medium cases, where both revisions comply with the instruction, similarity-based metrics outperform LLMs in identifying the best option.
We posit that these metrics can leverage information from the gold reference to assess the expected revision more effectively.

For hard cases, none of the metrics perform well, with all methods showing low alignment with human judgments, as the task becomes even more subjective. 
In such situations, the preference for one revision over another may depend largely on the original author's writing style and intent, making automatic evaluation difficult. 
However, \ppluie\ seems to be the best option for evaluating these hard cases, ensuring that the original meaning of the paragraph is preserved during the revision process and preventing the revision model from hallucinating.

Since LLMs struggle with correctness assessment, we further analyse this aspect in Appendix~\ref{apx:first_questions}, correlating results with preliminary human annotation questions. A full alignment breakdown including all metrics is available in Appendix~\ref{apx:all_correl}.

\subsection{Impact of Providing Gold Reference for LLM-as-a-Judge Approaches}

We examine whether providing the gold reference influences the performance of LLM-as-a-judge methods and find minimal impact. \textit{LLM-Choice} accuracy vary slightly from 0.564 to 0.563 when provided with the gold reference and \textit{LLM-Likert} from 0.436 to 0.457. Our findings contrast with~\citet{doostmohammadi-etal-2024-reliable}, who reported that in the absence of a reference, \gptIV\ exhibited weaker alignment with human judgments. This suggests that LLMs rely heavily on their own internal reasoning rather than direct comparisons to a gold-standard revision. More details in Appendix~\ref{apx:impact_gold}.

\section{Discussion and Conclusion}
\label{sec:discussion}
In this article, we identified the most reliable metrics to evaluate scientific text revision. The identification of those metrics will allow a better analyse- of existing text revisions strategies and designing new models for this task.

Our results suggest that LLMs-as-a-judge methods effectively assess whether a revision follows instructions but struggle to distinguish between two strong candidates and need to be completed by other metrics.
Traditional similarity metrics, while not designed to assess instruction-following, prove valuable in differentiating between valid revisions.
Their ability to compare revisions against a reference provides a tie-breaking mechanism when LLMs fail to make a clear distinction.

However, LLM-as-a-judge methods remain computationally expensive. To mitigate this, we recommend using a smaller, complementary set of metrics that strikes a balance between cost, interpretability, and alignment with human judgments. This subset could include a small LLM to evaluate instruction-following, \ppluie{} for meaning preservation, and similarity-based metrics like SARI and GLEU, which leverage information from the gold reference to help differentiate between more challenging cases.

\section{Limitations}

The primary limitation of this work is the size of the dataset, as we were limited by the size of the evaluation split of \dataset\ dataset and only had a limited number of volunteer researchers for manual annotation. As it need to be conducted with qualified annotators (researchers) our human resources are very limited, leading to a smaller double-annotated subset.
A larger amount of annotated data would enhance the reliability of our analysis, strengthening the claims we made in this paper.

Additionally, preference-based annotation is inherently subjective, as reflected in the Cohen’s Kappa scores in Table~\ref{tab:agreement}. For the ParaReval dataset, we first annotated a double-annotated subset and retained the annotations from researchers with the highest agreement. Those with the highest agreement scores continued the annotation process to enhance reliability. The choice of annotators (similar background) may also introduce a bias.

We were also limited by computational resources, as LLMs use a lot of energy.
Our priority was to ensure that our results were model-independent, so we conducted the same experience with different models but using only one prompt. However, it would have been preferable to also check the prompt-independent aspect~\citep{mizrahi-etal-2024-state}.

Finally, numerous methods and metrics have been proposed to evaluate tasks in text generation throughout the years. To keep our analysis clear, we considered a limited number of them. For the LLM-based approaches, a larger number of runs would have been preferable for \gptIV\ but due to cost issues, we had to limit it to one run per approach.

\section{Ethical Considerations}
\paragraph{Data availability}
All the data are from the \dataset\ corpus, the paragraphs are extracted from scientific articles collected on OpenReview where they fall under different "non-exclusive, perpetual, and royalty-free license"~\footnote{\href{https://openreview.net/legal/terms}{https://openreview.net/legal/terms}}.

\paragraph{Computational resources}
\begin{itemize}[topsep=.5em, itemsep=.1em,leftmargin=1em]
    \item To generate revisions with Co-edit and experiment with BERT-based metrics, we used a local GeForce RTX 2080 11Go GPU for approximately 12 hours.
    \item To use \ppluie\ and the different open LLMs to generate revisions and evaluation, we used V100 and A100 GPUs for a total of 249 hours on a supercomputer, equivalent to 0.009 tons of $CO_{2}$.
    \item To use \gptIVmini\ and \gptIV\ we spent 29.01\$ spent on GPT API credits.
\end{itemize}

\section*{Acknowledgments}

We thank Anas Belfathi, Maël Houbre, Trung Hieu Ngo, Xavier Pillet, Mohamed Reda Marzouk and Thomas Sebbag for their participation in human evaluation.

We thank Thomas Sebbag for proof-reading this paper.

This work was granted access to the HPC resources of IDRIS under the allocations 2023-AD011013901R1, 2024-AD011013901R2 and 2024-AD011014882R1 made by GENCI.

\bibliography{custom}

\appendix

\onecolumn

\section{Text Revision Prompt}
\label{apx:prompts_rev}
\input{acl_prompts/prompt_revision}

\section{Annotation Environment}
\label{apx:annot_env}
\begin{figure}[h]
  \includegraphics[width=\textwidth]{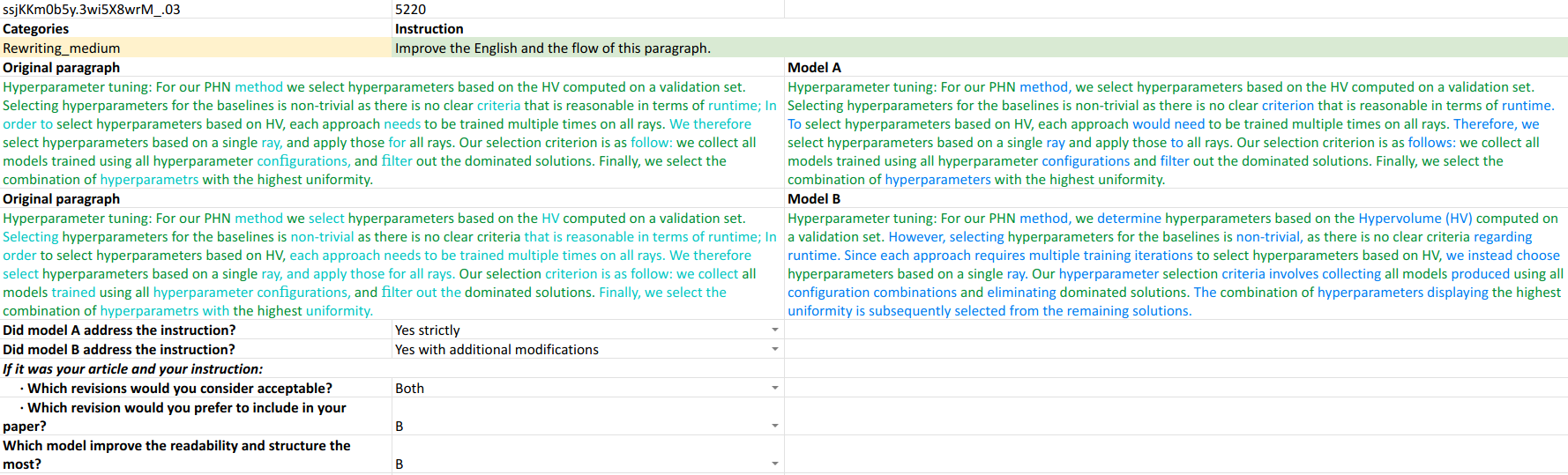}
  \caption{Screenshot of the annotation environment}
  \label{fig:screen}
\end{figure}

See Figure~\ref{fig:screen}

\newpage
\section{Human Pairwise Revision Model Preferences}
\label{apx:human_pref_matrix}
\begin{figure}[H]
\centering
  \includegraphics[width=0.7\textwidth]{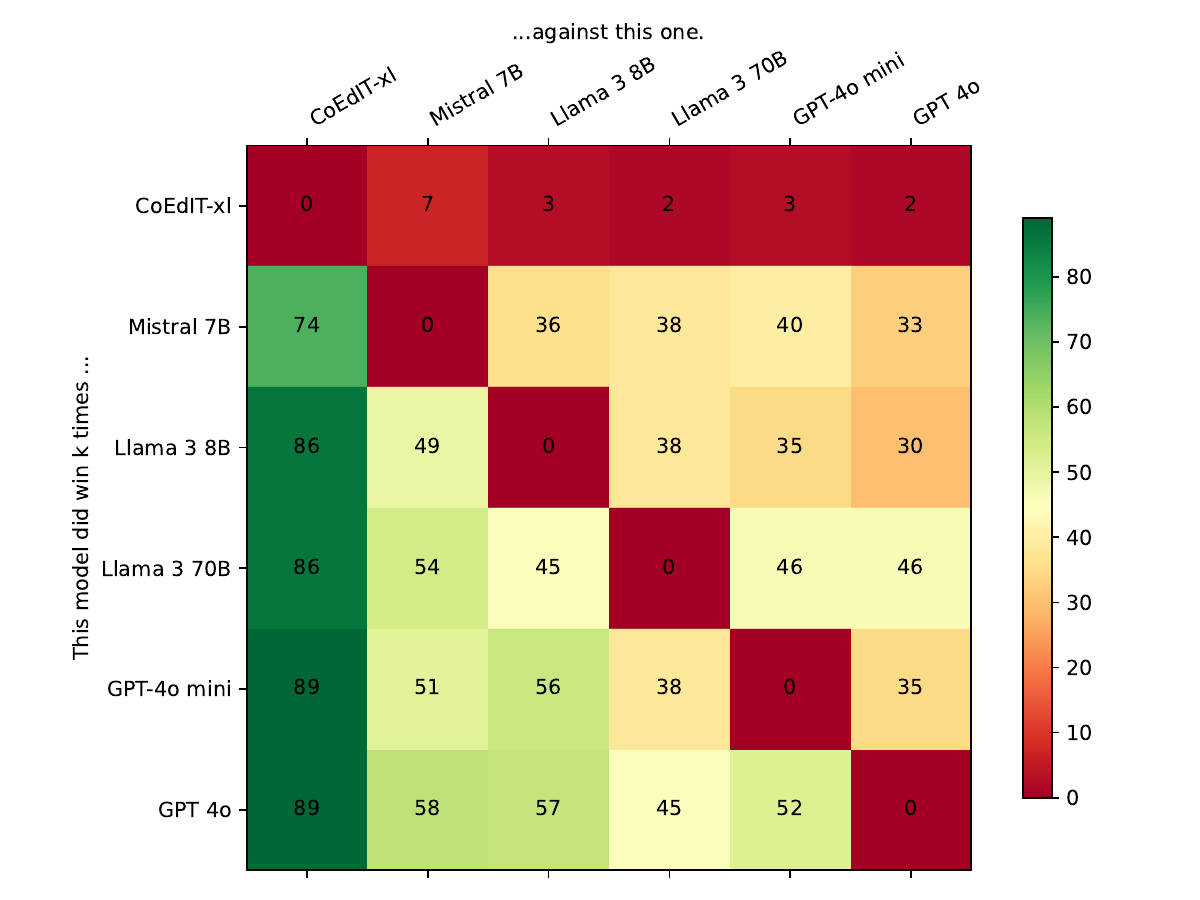}
  \caption{Pairwise comparison of human preferences on revision models.}
  \label{fig:human_pref_matrix}
\end{figure}

See Figure~\ref{fig:human_pref_matrix}

\newpage
\section{Relation between Metrics and Levenshtein Distance}
\label{apx:lev}
\begin{figure}[H]
  \includegraphics[width=\textwidth]{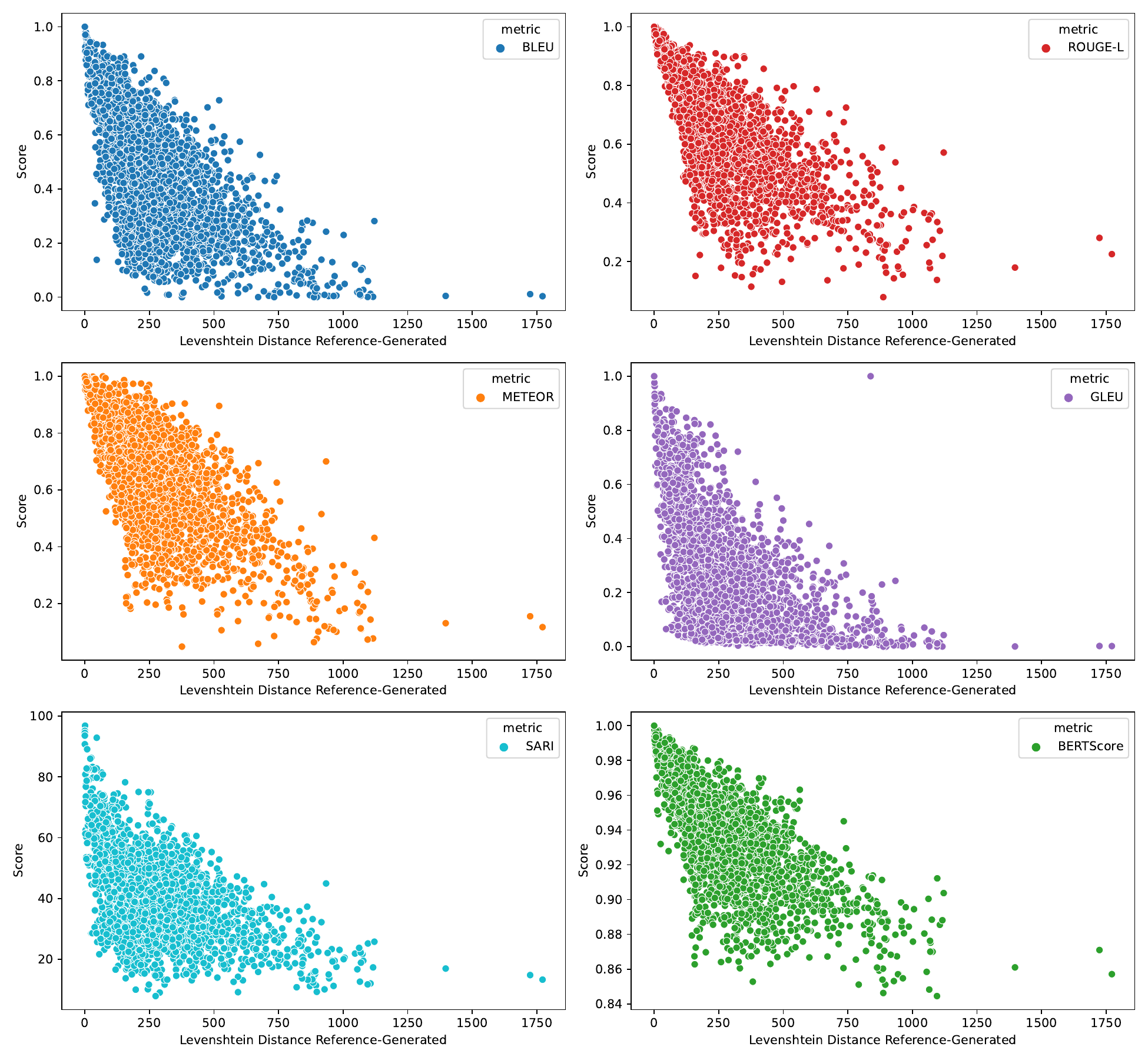}
  \caption{Distribution of similarity metrics scores based on Levenshtein distance between the reference and generated paragraph.}
  \label{fig:lev_1}
\end{figure}

\begin{figure}[H]
  \includegraphics[width=\textwidth]{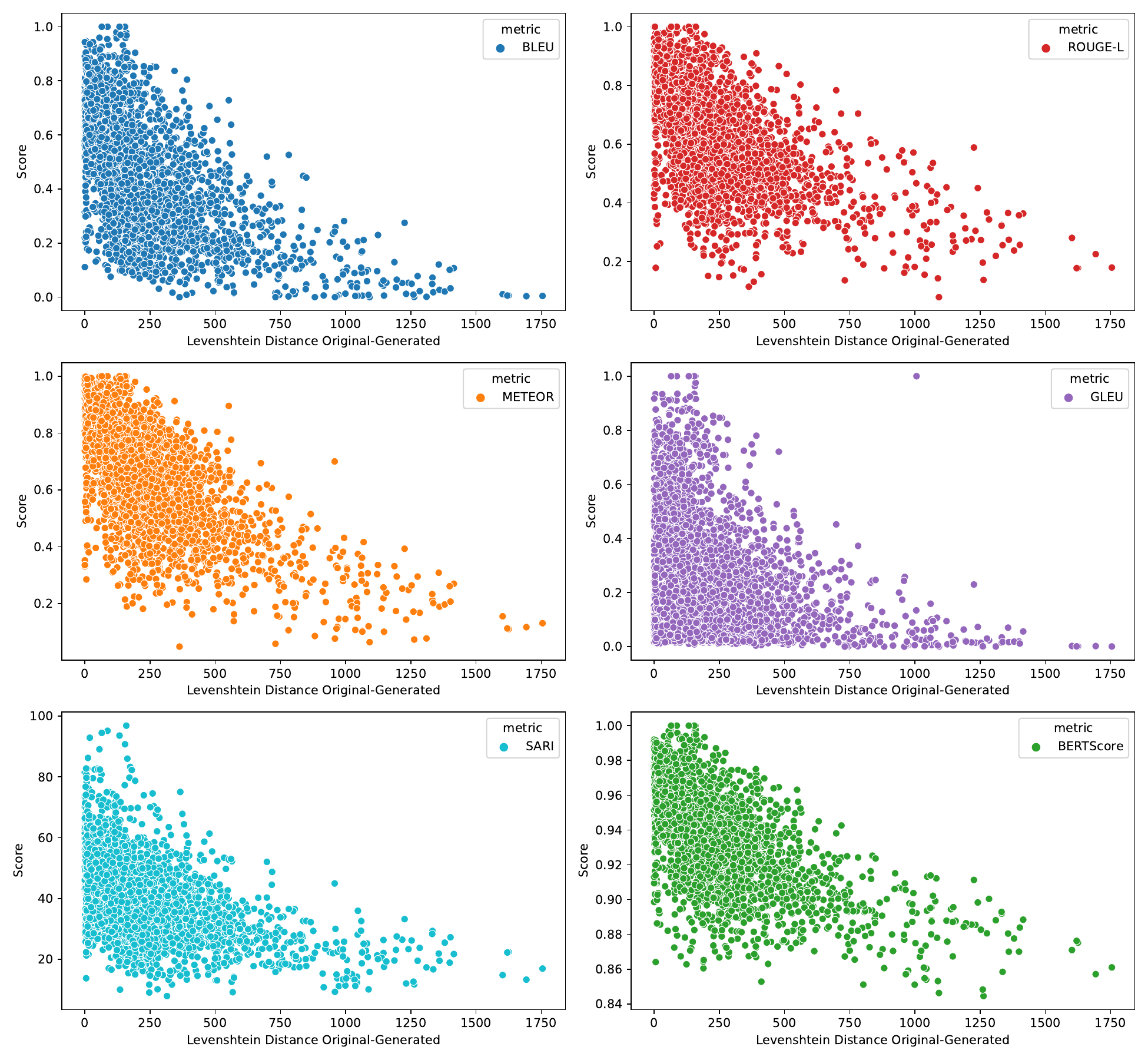}
  \caption{Distribution of similarity metrics scores based on Levenshtein distance between the original and generated paragraph.}
  \label{fig:lev_2}
\end{figure}

\begin{figure}[H]
  \includegraphics[width=\textwidth]{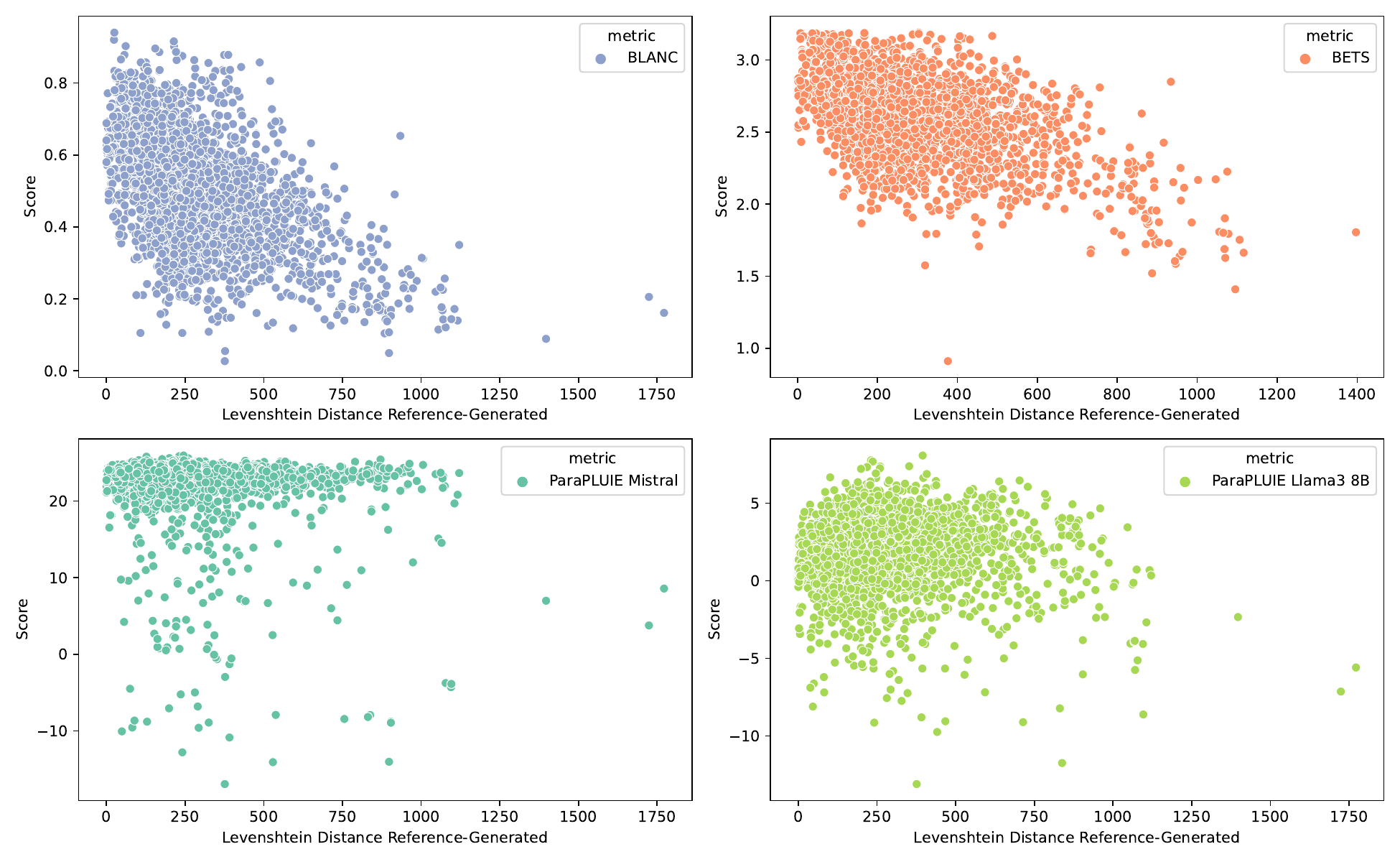}
  \caption{Distribution of alternative metrics scores based on Levenshtein distance between the reference and generated paragraph.}
  \label{fig:lev_3}
\end{figure}

\begin{figure}[H]
  \includegraphics[width=\textwidth]{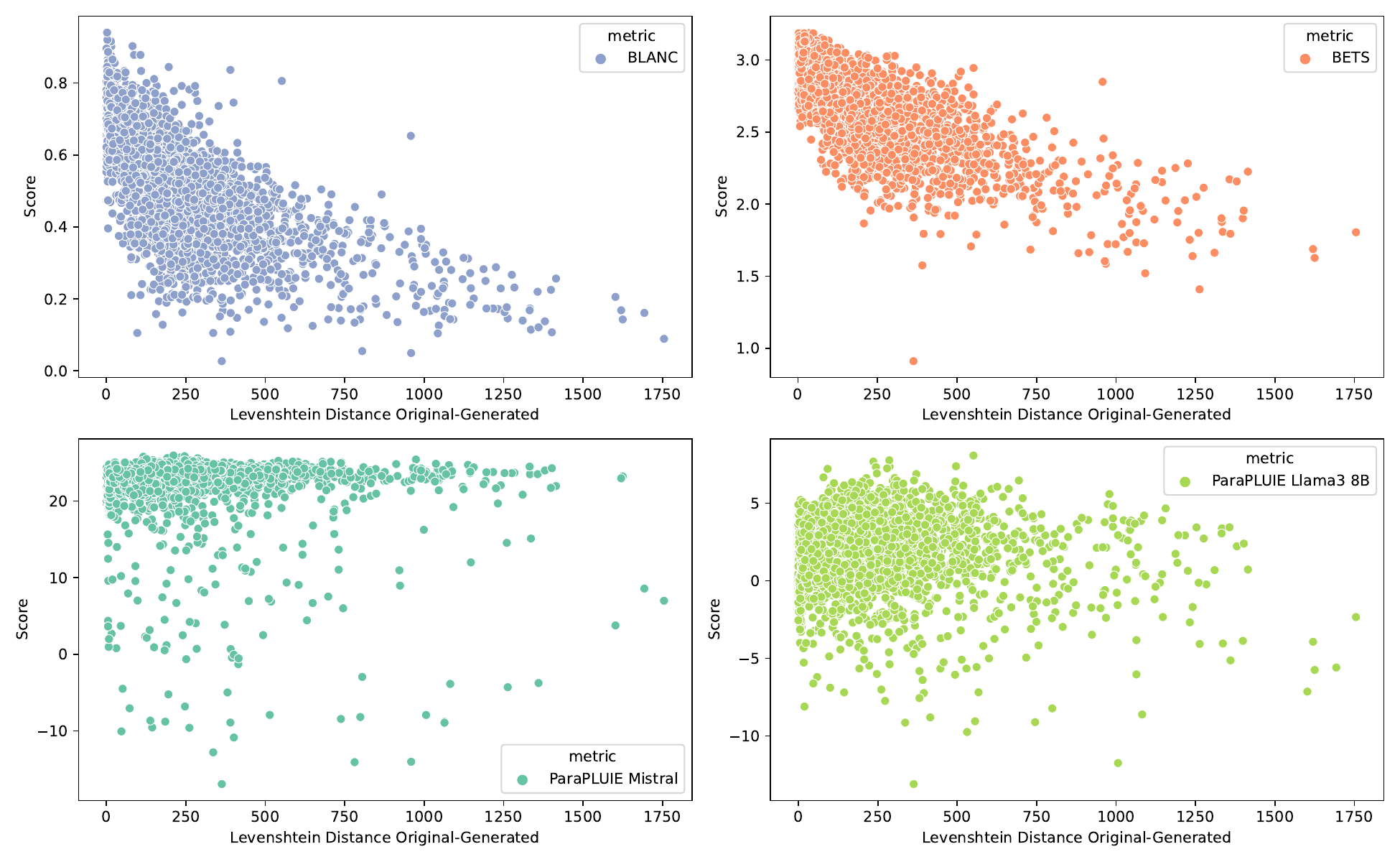}
  \caption{Distribution of alternative metrics scores based on Levenshtein distance between the original and generated paragraph.}
  \label{fig:lev_4}
\end{figure}

See Figures~\ref{fig:lev_1},~\ref{fig:lev_2},~\ref{fig:lev_3} and~\ref{fig:lev_4}.

\newpage


\newpage
\section{LLM-as-a-Judge Prompts}
\label{apx:prompts_judge}
Our prompts are inspired by the ones used in~\citet{doostmohammadi-etal-2024-reliable}.
See Figures~\ref{tab:gpt4prompt_choice}, \ref{tab:gpt4prompt_likert} and \ref{tab:gpt4prompt_likert_gold}for longer prompt sections.

\input{acl_prompts/LLM-Choice-System}

\input{acl_prompts/LLM-Choice-Gold-System}

\input{acl_prompts/Category-Questions-LLM-Choice}

\input{acl_prompts/LLM-Choice-User}

\input{acl_prompts/Category-Questions-LLM-Likert}

\input{acl_prompts/LLM-Likert-System}

\input{acl_prompts/LLM-Likert-User}

\input{acl_prompts/LLM-Likert-Gold-System}

\input{acl_prompts/LLM-Likert-Gold-User}

\newpage
\section{Bias of LLM Models on their own Revisions}
\label{apx:bias}

As we used several LLM for revision that we reused as judges, we check in Table~\ref{tab:llm-bias} if they are biased towards their own proposition. We don't observe such bias, and even notice that results tend to be consistent across judge models. However, as they all tend to favour \mistral\ we also computed the average edit distance between the original and generated revised texts for all revision models. As Mistral has the highest average, this could indicate an opposite bias as the one conveyed in similarity metrics: LLM-as-a-judge approaches tend to favour propositions with more important revisions.

\input{acl_tables/llm-bias}

\newpage
\section{Distribution of Extended Preference of each LLM Judge}
\label{sec:distrib_pref_llm_judge}
See Table~\ref{tab:distrib_pref_llm_judge}.

\input{acl_tables/distrib_pref_llm_judge}

\section{Distribution on Relatedness and Correctness for LLM-as-a-Judge Approaches}
 \label{apx:first_questions}
See Table~\ref{tab:first_questions}
\begin{table}[H]
  \centering
    \input{acl_tables/llm_additionnal_questions}
  \caption{Accuracy of LLM-Choice on the preliminary Relatedness and Correctness questions. For relatedness, in soft accuracy, we merge "Yes stricly" and "Yes with additional modifications" categories.}
  \label{tab:first_questions}
\end{table}

\section{Alignment of all Metrics}
\label{apx:all_correl}
\begin{figure}[H]
  \includegraphics[width=0.93\textwidth]{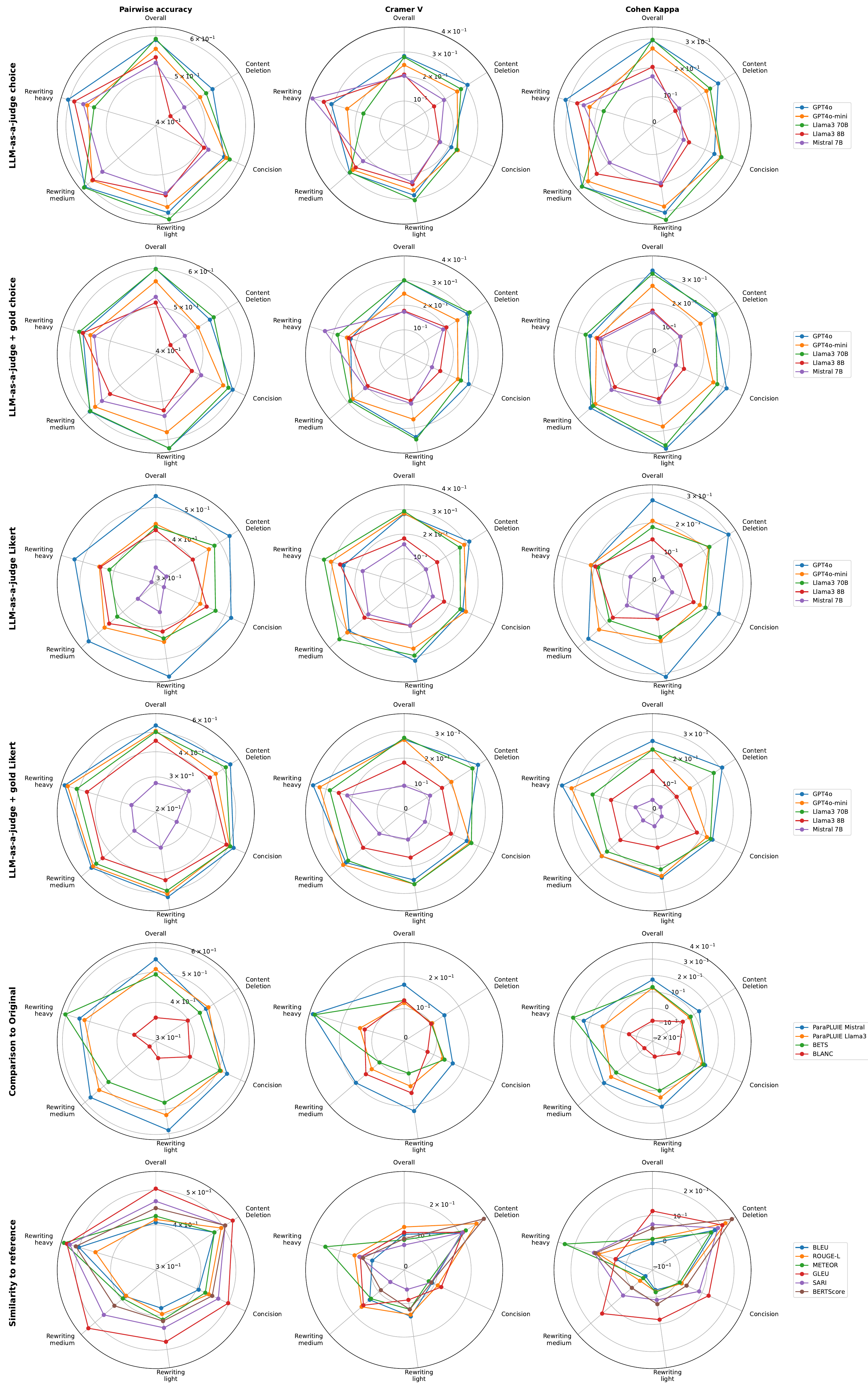}
  \caption{Alignment of metrics to human annotations by metric by type of metric and type of revision.}
  \label{fig:mega_alignement_cat}
\end{figure}

\begin{figure}[H]
  \includegraphics[width=0.93\textwidth]{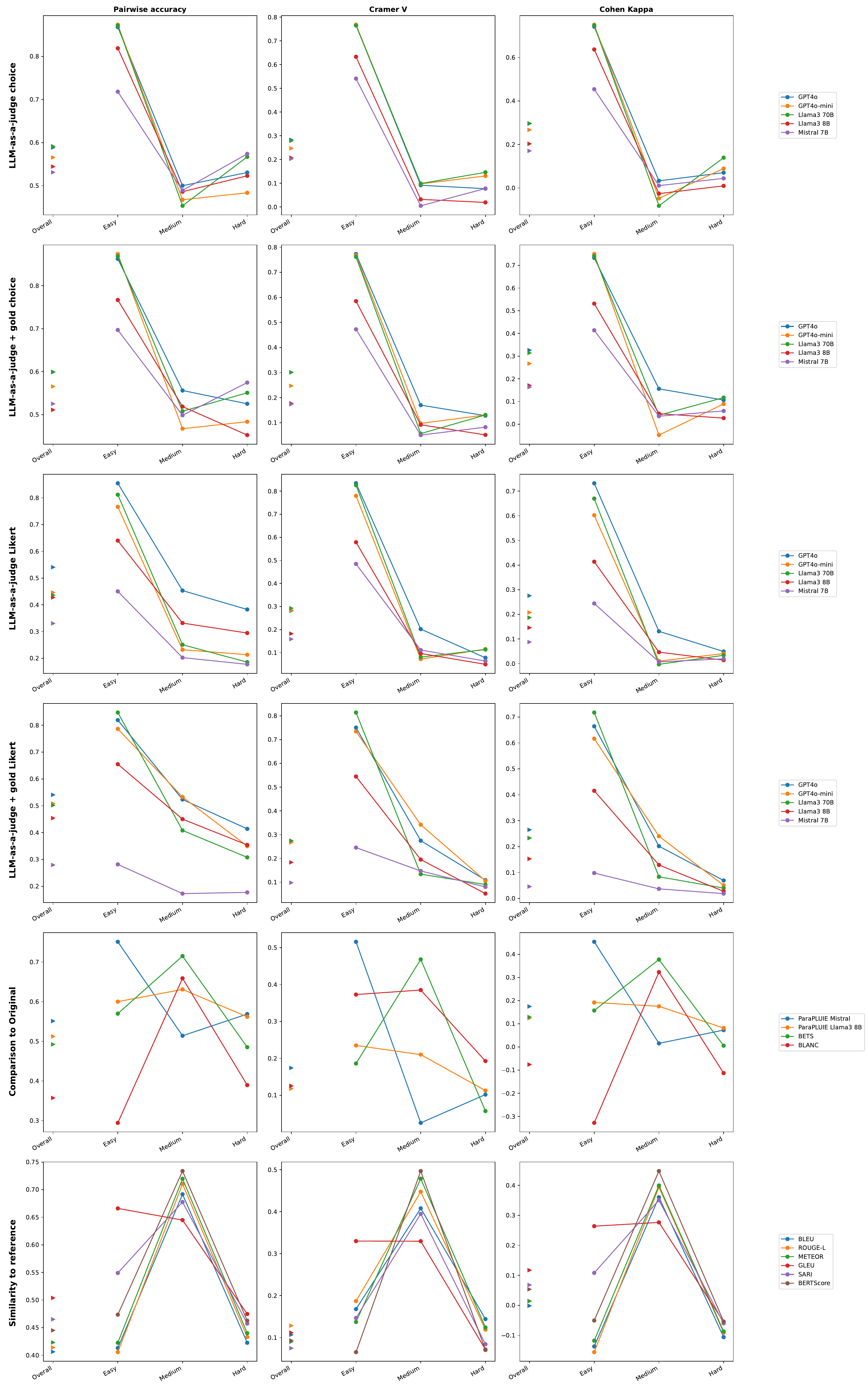}
  \caption{Alignment of metrics to human annotations by metric by types of difficulty}
  \label{fig:mega_alignement_diff}
\end{figure}

See Figures~\ref{fig:mega_alignement_cat} and \ref{fig:mega_alignement_diff}

\section{Impact of using Gold References on the Alignment of LLM-as-a-Judge Approaches}
\label{apx:impact_gold}
\begin{figure}[H]
  \includegraphics[width=\textwidth]{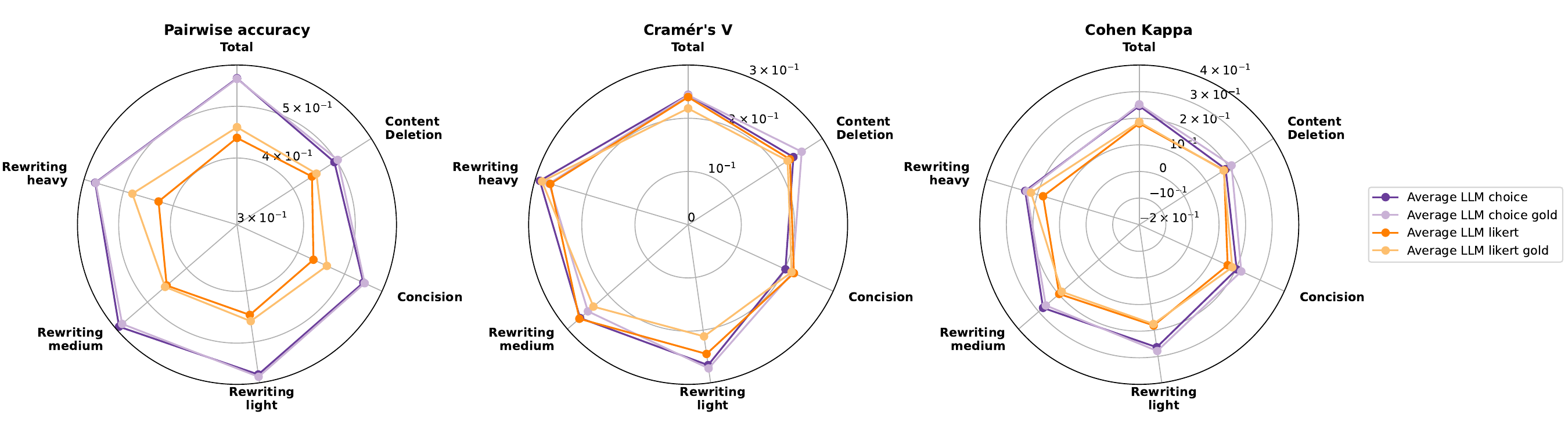}
  \caption{Alignment of LLM-as-a-judge approaches with human annotations by revision category}
  \label{fig:alignement_categories_gold}
\end{figure}

\begin{figure}[H]
  \includegraphics[width=\textwidth]{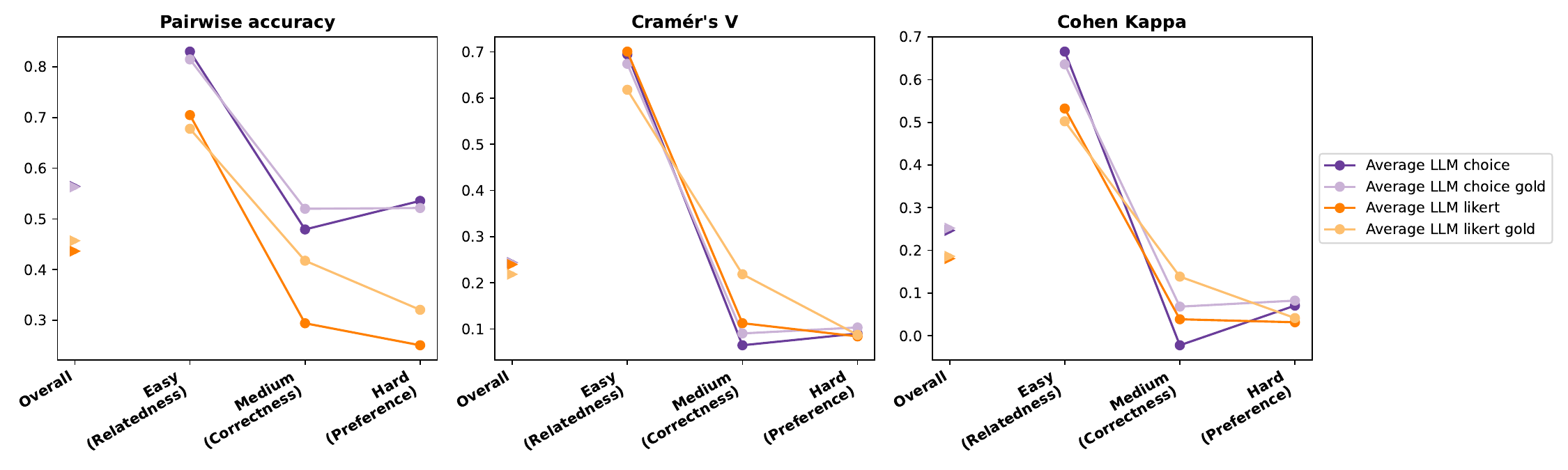}
  \caption{Alignment of LLM-as-a-judge approaches with human annotations, by difficulty. The triangles in the first column represent the agreement on the total dataset.}
  \label{fig:alignement_difficulty_gold}
\end{figure}

See Figures~\ref{fig:alignement_categories_gold} and~\ref{fig:alignement_difficulty_gold}.

\end{document}

%% file: acl_tables/taxonomy-pararev.tex
\begin{tabular}{lll}
    \hline
    \textbf{Type} & & \textbf{Description} \\
    \hline
                   & Light         &   Minor changes in word choice\\
                   & & or phrasing.\\
    Rewriting      & Medium        &  Complete rephrasing of sentences\\
                   & & within the paragraph.  \\
                   & Heavy         &  Significant rephrasing, affecting\\
                   & & at least half of the paragraph.\\\hline
    Concision      &               &Same idea, stated more briefly by\\
                   & &removing unnecessary details.\\\hline
         Content          & Deletion      &   Modification of content through\\
                   & &the deletion of an idea.  \\\hline
  \end{tabular}

%% file: acl_tables/agreement.tex
 \begin{tabular}{l | cr}
    \toprule
    \textbf{Question} & \textbf{$\kappa$} & \textbf{Agreement} \\
    \midrule
    Relatedness & 0.54 & Moderate\\
    Correctness & 0.55 & Moderate \\
    Preference & 0.33 & Fair \\
    Concision & 0.22 & Fair\\
    Rewriting light & 0.41 & Moderate \\
    Rewriting medium & 0.48 & Moderate \\
    \bottomrule
  \end{tabular}

%% file: acl_tables/results_all.tex
 \begin{tabular}{l| rrrrrr|rrr}
    \toprule
    Rev. Model& \textbf{BLEU} &\textbf{ROUGE-L} & \textbf{METEOR} & \textbf{GLEU}&\textbf{SARI} & \textbf{Bertscore} & \textbf{BLANC} & \textbf{BETS} & \textbf{ParaPLUIE} \\
    \midrule
    \textcolor{violet}{no edits} | \textcolor{brown}{gold} & \textcolor{violet}{\textbf{66.00}} & \textcolor{violet}{\textbf{78.30}} & \textcolor{violet}{\textbf{83.80}} & \textcolor{violet}{25.78} & \textcolor{violet}{\textbf{60.63}} & \textcolor{violet}{\textbf{95.95}} & \textcolor{brown}{\textit{\underline{55.21}}} & \textcolor{brown}{\textit{2.461}}  & \textcolor{brown}{\textit{20.93}}\\\cline{8-10}
    CoEdIT-XL & 50.24 & 67.46 & 66.66 & 23.84 & 39.60 & 93.90 &  \textbf{58.96} & 1.554 & 19.35\\
    Mistral-7B & 27.77 & 50.79 & 54.02 & 15.38 & 31.63 & 92.14 & 41.59 & \underline{2.491} & \textbf{23.02} \\
    Llama-3-8B & 41.66 & 62.07 & 62.00 & 25.78 & 39.33 & 93.53 & 49.09 & 2.364 & 22.67\\
    Llama-3-70B & 46.78 & 65.61 & 67.20 & 30.31 & 42.74 & 93.90 & 52.27 & 2.386 & 22.58 \\
    GPT4o-mini & \underline{51.68} & \underline{69.54} & \underline{72.70} & \textbf{32.67} &\underline{45.06} & \underline{94.80} & \underline{54.89} & \textbf{2.497} & 22.74\\
    GPT4o & 49.34 & 68.20 & 69.88 & \underline{31.35} &43.54 & 94.45 & 53.62 & 2.454 & \underline{22.86}\\
    \bottomrule
  \end{tabular}  

%% file: acl_tables/results-llm.tex
   \begin{tabular}{l | rrrr}
   \toprule
     & \multicolumn{2}{c}{\textbf{LLM-choice}} & \multicolumn{2}{c}{\textbf{LLM-likert}}  \\
    Rev. Model &\textbf{base}  & \textbf{w. gold} & \textbf{base} & \textbf{w. gold}  \\
    \midrule
    CoEdIT-XL & 6.15  & 8.82 & 3.385 & 3.487 \\ 
    Mistral-7B & \textbf{64.95} & 56.73 & \textbf{4.816} & \textbf{4.497} \\ 
    Llama-3-8B & 58.88 & \underline{57.08} & \underline{4.789} & 4.472\\ 
    Llama-3-70B & \underline{59.34} & \textbf{60.20} & 4.784 & \underline{4.479}\\ 
    GPT4o-mini & 50.52 & 52.33 & 4.750 & 4.447 \\ 
    GPT4o & 50.85 & 51.41 & 4.743 & 4.443 \\ 
    \bottomrule
  \end{tabular} 

%% file: acl_tables/overall_results.tex
\begin{tabular}{ l| rrr}
    \toprule
    \textbf{Judge} & \textbf{Pair acc.} &\textbf{$V$} & \textbf{$\kappa$} \\
    \midrule
    Avg. LLM choice & \textbf{0.564} & \textbf{0.244} & \textbf{0.247}\\
    
    Avg. LLM likert & 0.436 & 0.240 & 0.181\\
    
    ParaPLUIE & \underline{0.551} & \underline{0.241} & \underline{0.218}\\
    
    BETS & 0.492 & 0.152 & 0.127\\
    
    BLANC & 0.357 & 0.117 & -0.080\\
    BERTScore & 0.445 & 0.161 & 0.034 \\
    SARI & 0.465 & 0.183 & 0.071\\
    GLEU & 0.504 & 0.193 & 0.138\\
    ROUGE-L & 0.414 & 0.179 & -0.013\\
    \textit{Random} & \textit{0.334} & \textit{0.027} & \textit{0.003} \\
    \bottomrule
\end{tabular}  

%% file: acl_prompts/prompt_revision.tex
\begin{codelisting} {Text revision prompt messages}
system_message= """You are a writing assistant specialised in academic writing. 
Your task is to revise the original paragraph from a research paper draft that will be given according to the author's instruction. The input will follow the pattern ' <author_instruction> : "<Original_paragraph>" '.      	 
Please answer only by "Revised paragraph: <revised_version_of_the_paragraph>". Please limit your modifications only to what is requested in the author's instruction. Do not make any other modifications to the rest of the paragraph."""
     
user_message= """(*@\textcolor{blue}{\{instruct\}}@*) : \"(*@\textcolor{blue}{\{parag\}}@*)\" """
\end{codelisting}

%% file: acl_prompts/LLM-Choice-System.tex
\begin{codelisting} {LLM Choice System message without Gold}
You are an evaluator of academic writing on the task of text revision. 
In this task, two revision models have been provided with the original paragraph written for a scientific article and a revision instruction on how to revise the paragraph.
You will be given the proposition from the two different models and several questions to determine the quality of those propositions and identify the best one. In your answer please only provide the answers to the questions.
\end{codelisting}

%% file: acl_prompts/LLM-Choice-Gold-System.tex
\begin{codelisting} {LLM Choice System message with Gold}
You are an evaluator of academic writing on the task of text revision. 
In this task, two revisions models have been provided with the original paragraph written for a scientific article and a revision instruction on how to revise the paragraph.
You will be given the proposition from the two different models and several questions to determine the quality of those proposition and identify the best one. (*@\textbf{To help you in this task you will also be given the gold paragraph which is the version revised by the author themselves.}@*)
In your answer please only provide the answers to the questions.
\end{codelisting}

%% file: acl_prompts/Category-Questions-LLM-Choice.tex
\begin{codelisting} {Category Questions for LLM Choice with and without gold}
Rewriting_light:"""Which model improves the academic style and English the most?"""

Rewriting_medium:"""Which model improves the readability and structure the most?"""
                            
Rewriting_heavy:"""Which model improves the readability and clarity the most?"""
                            
Concision:"""Which model manages the most to give a shorter version while keeping all the important ideas?"""
    
\end{codelisting}

%% file: acl_prompts/LLM-Choice-User.tex
\begin{figure}[H]
\begin{codelisting} {LLM Choice User message with and without Gold}
[BEGIN DATA]
***
[Original paragraph]: \"(*@\textcolor{blue}{\{original\_paragraph\}}@*)\"
***
[Revision instruction]: \"(*@\textcolor{blue}{\{instruction\}}@*)\"
***
[Model A]: \"(*@\textcolor{blue}{\{modelA\_generated\_revised\_paragraph\}}@*)\"
***
[Model B]: \"(*@\textcolor{blue}{\{modelB\_generated\_revised\_paragraph\}}@*)\"
***
[END DATA]

1. Did model A address the instruction? Answer "Yes strictly", "Yes with additional modifications" or "No":
 - Yes strictly : The model proposition matches what is required in the instruction. Here, the quality of the revision does not matter.
 - Yes with additional modifications : The model proposed additional modifications to the one required in the instruction. But some of the modification address the needs stated in the instruction.
- No : The model proposition does not match the instruction.

2. Did model B address the instruction? (Answer "Yes strictly", "Yes with additional modifications" or "No")

3. Is model A revision acceptable? Answer "Yes" or "No". Answer "Yes" if the model made a good quality revision proposition that should replace the original paragraph in the scientific article.

4. Is model B revision acceptable? (Answer "Yes" or "No")

5. Which model proposed the best revision? (Answer preferably "A" or "B", you can answer "Both" if it is really a tie. Answer "None" if you answered "No" to question 3 and 4.)"""
<Additional category questions depending on the revision intention labels of the instance>

"""For all questions, you do not need to explain the reason.

Your response must be RFC8259 compliant JSON following this schema: 
{{"1": str, "2": str , "3": str , "4": str , "5": str """
< """ "6": str """ and """, "7": str """ can be added depending on the number of labels of the instance.>
"""}}
\end{codelisting}
\caption{User message for prompting LLM Choice with and without Gold}
\label{tab:gpt4prompt_choice}
\end{figure}

%% file: acl_prompts/Category-Questions-LLM-Likert.tex
\begin{codelisting} {Category Questions for LLM Likert with and without gold}
Rewriting_light:"""The academic style and english has been improved."""

Rewriting_medium:"""The readability and structure has been improved."""
                            
Rewriting_heavy:"""The paragraph has been rewritten in a more well organized and clear version, fitting the academic style."""
                            
Concision:"""The generated revision is a shorter version that kept all the important ideas."""
    
\end{codelisting}

%% file: acl_prompts/LLM-Likert-System.tex
\begin{codelisting} {LLM Likert System message without Gold}
You are an evaluator of academic writing on the task of text revision. 
In this task, (*@\textbf{a revision model}@*) have been provided with the original paragraph written for a scientific article and a revision instruction on how to revise the paragraph.
You will be given the proposition from the (*@\textbf{revision model}@*) and several (*@\textbf{affirmations}@*) to determine the quality of (*@\textbf{this}@*) proposition(*@\textbf{.}@*)
(*@\textbf{You will answer each affirmation with a grade (int) from 1 to 5 as following: 1 = Strongly disagree , 2 = Disagree , 3 = Neutral ,4 = Agree , 5 = Strongly agree }@*)
In your answer please only provide the answers to the (*@\textbf{affirmations}@*).
\end{codelisting}

%% file: acl_prompts/LLM-Likert-User.tex
\begin{figure}[H]
\begin{codelisting} {LLM Likert User message without Gold}

[BEGIN DATA]
***
[Original paragraph]: \"(*@\textcolor{blue}{\{original\_paragraph\}}@*)\"
***
[Revision instruction]: \"(*@\textcolor{blue}{\{instruction\}}@*)\"
***
(*@\textbf{[Model proposed revision]: }@*)\"(*@\textbf{\textcolor{blue}{\{model\_generated\_revised\_paragraph\}}}@*)\"
***
[END DATA]

(*@\textbf{1. Relatedness: The generated revision correctly addressed the instruction.}@*)

(*@\textbf{2. Correctness: The generated revision is better than original version in my opinion.}@*)"""
<Additional category questions depending on the revision intention labels of the instance>

"""For all questions, you do not need to explain the reason.

Your response must be RFC8259 compliant JSON following this schema: 
{{"1": str, "2": str , "3": str , "4": str , "5": str """
< """ "6": str """ and """, "7": str """ can be added depending on the number of labels of the instance.>
"""}}
\end{codelisting}
\caption{User message for prompting LLM Likert without Gold}
\label{tab:gpt4prompt_likert}
\end{figure}

%% file: acl_prompts/LLM-Likert-Gold-System.tex
\begin{codelisting} {LLM Likert System message with Gold}
You are an evaluator of academic writing on the task of text revision. 
In this task, a revision model have been provided with the original paragraph written for a scientific article and a revision instruction on how to revise the paragraph.
You will be given the proposition from the revision model and several affirmations to determine the quality of this proposition.
You will answer each affirmation with a grade (int) from 1 to 5 as following: 1 = Strongly disagree , 2 = Disagree , 3 = Neutral ,4 = Agree , 5 = Strongly agree 
(*@\textbf{To help you in this task you will also be given the gold paragraph which is the version revised by the author themselves.}@*)
In your answer please only provide the answers to the affirmations.
\end{codelisting}

%% file: acl_prompts/LLM-Likert-Gold-User.tex
\begin{figure}[H]
\begin{codelisting} {LLM Likert User message with Gold}

[BEGIN DATA]
***
[Original paragraph]: \"(*@\textcolor{blue}{\{original\_paragraph\}}@*)\"
***
[Revision instruction]: \"(*@\textcolor{blue}{\{instruction\}}@*)\"
***
[Model proposed revision]: \"(*@\textcolor{blue}{\{model\_generated\_revised\_paragraph\}}@*)\"
***
(*@\textbf{[Gold revised paragraph]: }@*)\"(*@\textcolor{blue}{\textbf{\{gold\}}}@*)\"
(*@\textbf{***}@*)
[END DATA]

(*@\textbf{1. Gold similarity: The generated revision is similar to gold revision.}@*)

(*@\textbf{2.}@*) Relatedness: The generated revision correctly addressed the instruction.

(*@\textbf{3.}@*) Correctness: The generated revision is better than original version in my opinion."""
<Additional category questions depending on the revision intention labels of the instance>

"""For all questions, you do not need to explain the reason.

Your response must be RFC8259 compliant JSON following this schema: 
{{"1": str, "2": str , "3": str , "4": str , "5": str """
< """ "6": str """ and """, "7": str """ can be added depending on the number of labels of the instance.>
"""}}
\end{codelisting}
\caption{User message for prompting LLM Likert with Gold}
\label{tab:gpt4prompt_likert_gold}
\end{figure}

%% file: acl_tables/llm-bias.tex
\begin{table}[H]
\begin{adjustwidth}{-0.9cm}{}
  \centering
\resizebox{\linewidth}{!}{
\begin{tabular}{c| r  r r r r r}
\hline
    \textbf{Judge$\downarrow$/Revision model$\rightarrow$}& \textbf{CoEdIT} & \textbf{Mistral 7B} & \textbf{Llama 3 8B} & \textbf{Llama 3 70B} & \textbf{GPT-4o mini} & \textbf{GPT 4o}\\\hline
    Human & 3.29 & 42.83 & 46.12 & \underline{53.68} & 52.13 & \textbf{58.33} \\\hline
   \mistral\ Choice & 21.58 & \textbf{60.92} & \underline{56.20} & 55.17 & 50.65 & 52.65\\
   \LlamaH\ Choice & 3.94 & \textbf{64.66} & 58.98 &  \underline{62.02} & 58.08 & 52.07\\
   \LlamaS\ Choice & 1.49 & \textbf{73.00} & \underline{61.76} & 59.23 & 46.06 & 48.13 \\
    \gptIVmini\ Choice& 1.81 & \textbf{66.86} & \underline{58.53} & 57.30 & 47.61 & 48.90\\
    \gptIV\ Choice& 1.94 & \underline{59.30} & 58.91 & \textbf{62.98} & 50.19 & 52.52\\ \hline
   \mistral\ Gold Choice & 23.64 & 54.65 & \underline{56.33} & \textbf{56.52} & 51.16 & 54.07\\
   \LlamaH\ Gold Choice & 14.67 & 51.62 & \underline{59.24} & \textbf{60.06} & 53.68 & 50.74\\
   \LlamaS\ Gold Choice & 1.49 & \textbf{64.47} & 58.85 & \underline{61.95} & 51.10 & 48.19\\
    \gptIVmini\ Gold Choice & 2.00 & \textbf{62.73} & 58.08 & \underline{59.50} & 51.61 & 51.16\\
    \gptIV\ Gold Choice& 2.33 & 50.19 & 52.91 & \textbf{62.98} & \underline{54.07} & 52.91\\\hline
    \mistral\ Likert  & 5.04 & \underline{25.65} & 24.55 & 23.84 & \textbf{25.97} & 23.25\\
    \LlamaH\ Likert& 6.98 & 38.11 & \textbf{39.99} & 36.24 & \underline{38.56} & 36.05\\
    \LlamaS\ Likert& 2.26 & \textbf{40.38} & \underline{37.40} & 32.69 & 29.20 & 30.10 \\
    \gptIVmini\ Likert & 1.87 & \textbf{36.95} & \underline{33.85} & 31.98 & 31.52 & 29.39 \\
   \gptIV\ Likert & 1.36 & 41.28 & \underline{47.87} & \textbf{49.61} & 47.67 & 46.90\\\hline
   \mistral\ Likert Gold & 11.24 & 18.41 & 17.89 & \underline{19.64} & 19.05 & \textbf{19.77}\\
    \LlamaH\ Likert Gold& 10.92 & \textbf{46.25} & 41.15 & \underline{42.96} & 42.18 & 39.60\\
   \LlamaS\ Likert Gold & 2.45 & \underline{46.19} & \textbf{47.22} & 44.51 & 39.28 & 38.76\\
   \gptIVmini\ Likert Gold & 2.58 & 44.06 & 43.99 & \underline{45.41} & \textbf{46.51} & 44.19 \\
   \gptIV\ Likert Gold  & 3.88 & 42.25 & \underline{49.22} & \underline{49.22} & 48.45 & \textbf{51.16}\\\hline
   \ppluie\ Mistral & 17.83 & \textbf{72.48} & \underline{56.20} & 52.91 & 47.67 & 50.39\\
   \ppluie\ Llama3 8B & 20.35 & \textbf{70.74} & 52.33 & \underline{54.84} & 52.33 & 46.90\\\hline\hline
   Edit distance (Original-Generated) &190.82 &\textbf{342.69} & \underline{270.47} & 234.95 & 175.02 & 197.36\\\hline
  \end{tabular} 
  }
  \caption{Distribution of extended strict preference of each LLM judge by revision model}
  \label{tab:llm-bias}
  \end{adjustwidth}
\end{table}

%% file: acl_tables/distrib_pref_llm_judge.tex
\begin{table}[H]
  \centering
\begin{tabular}{c|c c c}
\hline
    \textbf{Judge$\downarrow$/Choice$\rightarrow$}& \textbf{Tie} & \textbf{A} & \textbf{B}\\\hline
    Human & 14.53 & 44.25 & 41.21 \\\hline
   \mistral\ Choice & 0.95 & 82.43 & 16.63\\
   \LlamaH\ Choice & 0.08 & 53.92 & 46.00\\
   \LlamaS\ Choice & 03.45 & 52.97 & 43.58\\
    \gptIVmini\ Choice& 06.33 & 39.28 & 54.39\\
    \gptIV\ Choice& 04.72 & 53.62 & 41.67\\ \hline
   \mistral\ Gold Choice & 01.21 & 77.76 & 21.04\\
   \LlamaH\ Gold Choice & 03.33 & 23.79 & 72.85 \\
   \LlamaS\ Gold Choice & 04.65 & 51.14 & 44.21\\
    \gptIVmini\ Gold Choice & 4.98 & 39.27 & 55.75 \\
    \gptIV\ Gold Choice& 08.21 & 49.94 & 41.86 \\\hline
    \mistral\ Likert& 57.24 & 22.37 & 20.39 \\
    \LlamaH\ Likert& 34.69 & 34.28 & 31.03\\
    \LlamaS\ Likert& 42.66 & 28.71 & 28.64 \\
    \gptIVmini\ Likert & 44.81 & 27.99 & 27.2  \\
   \gptIV\ Likert & 21.77 & 39.02 & 39.21 \\\hline
   \mistral\ Likert Gold & 64.66 & 17.29 & 18.05\\
    \LlamaH\ Likert Gold& 25.65 & 38.20 & 36.16\\
   \LlamaS\ Likert Gold & 27.20 & 35.44 & 37.36  \\
   \gptIVmini\ Likert Gold & 24.42 & 37.96 & 37.62 \\
   \gptIV\ Likert Gold & 18.60 & 41.54 & 39.86  \\
   \hline
   \end{tabular}  
\caption{Distribution of extended preference of each LLM judge}
  \label{tab:distrib_pref_llm_judge}
\end{table}

%% file: acl_tables/llm_additionnal_questions.tex
 \begin{tabular}{l| rr|r}
 \toprule
 & \textbf{Relatedness} & \textbf{Relatedness} & \textbf{Correctness}\\
\textbf{Model} & \textbf{Strict Acc.} & \textbf{Soft Acc.} & \textbf{Acc.}\\
\midrule
gpt-4o &  \textbf{67.03} & 84.69 & \underline{76.78}\\
gpt-4o-gold & 62.99 & 84.66 &  \textbf{76.94}\\
gpt-4o-mini & 62.26 & \textbf{85.30} & 76.04\\
gpt-4o-mini-gold & 58.95 & \underline{85.14} & 75.91\\
llama3-70b  & \underline{66.66} & 82.11 & 75.96\\
llama3-70b-gold  & 66.48 & 82.37 & 76.10\\
llama3-8b & 45.41 & 80.16 & 72.30\\
llama3-8b-gold &  40.55 & 77.70 & 61.93\\
mistral-gold & 58.78 & 75.19 & 62.37\\
mistral & 58.98 & 75.27 & 62.58\\
\bottomrule
\end{tabular}